\documentclass{article}

\usepackage[final,nonatbib]{includes/neurips_2021}

\usepackage[ruled,vlined]{algorithm2e}
\usepackage{algorithmic}
\usepackage{multirow}

\usepackage{amsthm}
\usepackage{amsmath}
\usepackage{amssymb}
\usepackage{bbold}
\usepackage{float}
\usepackage{graphicx}
\usepackage{listings}
\usepackage{xcolor}
\usepackage{pifont}
\usepackage{enumitem}
\usepackage{xspace}

\usepackage{hyperref}
\usepackage[capitalise]{cleveref}

\usepackage{subfigure}

\newcommand{\ourtitle}{Uniform Sampling over Episode Difficulty}
\newcommand{\ess}{\text{ESS}}
\newcommand{\uniform}{\textsc{Uniform}\xspace}
\newcommand{\easy}{\textsc{Easy}\xspace}
\newcommand{\hard}{\textsc{Hard}\xspace}
\newcommand{\curriculum}{\textsc{Curriculum}\xspace}
\def \mimg {Mini-ImageNet\xspace}
\def \timg {Tiered-ImageNet\xspace}
\def \cnn {conv($64$)$_{4}$\xspace}
\def \resnet {ResNet-$12$\xspace}
\def \osfw {$1$-shot $5$-way\xspace}
\def \fsfw {$5$-shot $5$-way\xspace}

\setlist{nolistsep}

\newcommand{\ie}{\emph{i.e.}\xspace}

\definecolor{light-gray}{gray}{0.65}
\definecolor{maureen-blue}{rgb}{0.2235, 0.4157, 0.6941}
\definecolor{maureen-orange}{rgb}{0.854902, 0.4863, 0.1882353}
\definecolor{maureen-green}{rgb}{0.243137, 0.588235, 0.317647}
\definecolor{maureen-red}{rgb}{0.811764, 0.145098, 0.160784}
\definecolor{maureen-black}{rgb}{0.325490, 0.317647, 0.329411}
\definecolor{maureen-purple}{rgb}{0.419607, 0.298039, 0.603921}
\definecolor{maureen-cardinal}{rgb}{0.572549, 0.141176, 0.156862}
\definecolor{maureen-gold}{rgb}{0.580392, 0.545098, 0.239215}
\definecolor{seb-blue}{rgb}{0.174706, 0.654902, 0.905882}

\newcommand{\brackets}[1]{\left[ #1 \right]}

\DeclareMathOperator*{\E}{\mathop{\mathrm{E}}}
\newcommand{\Exp}[2][]{\E_{#1}\brackets{#2}}

\newcommand{\eat}[1]{}

\usepackage[utf8]{inputenc} 
\usepackage[T1]{fontenc}    
\usepackage{hyperref}       
\usepackage{url}            
\usepackage{booktabs}       
\usepackage{amsfonts}       
\usepackage{nicefrac}       
\usepackage{microtype}      
\usepackage{xcolor}         
\usepackage{wrapfig} 
\usepackage{caption} 
\usepackage[numbers]{natbib}

\title{\ourtitle}

\author{
    S\'{e}bastien M. R. Arnold$^{1}$\thanks{Equal contributions; work done while at Amazon Web Services.}$\:\:$,
    Guneet S. Dhillon$^{2}$\footnotemark[1]$\:\:$,
    Avinash Ravichandran$^{3}$,
    Stefano Soatto$^{3,4}$
    \\
    $^{1}$University of Southern California,
    $^{2}$University of Oxford,
    \\
    $^{3}$Amazon Web Services,
    $^{4}$University of California, Los Angeles
    \\
    \texttt{seb.arnold@usc.edu, guneet.dhillon@stats.ox.ac.uk,}
    \\
    \texttt{ravinash@amazon.com, soattos@amazon.com}
}

\begin{document}

\maketitle

\begin{abstract}
    Episodic training is a core ingredient of few-shot learning to train models on tasks with limited labelled data.
    Despite its success, episodic training remains largely understudied, prompting us to ask the question: what is the best way to sample episodes? 
    In this paper, we first propose a method to approximate episode sampling distributions based on their difficulty.
    Building on this method, we perform an extensive analysis and find that sampling uniformly over episode difficulty outperforms other sampling schemes, including curriculum and easy-/hard-mining.
    As the proposed sampling method is algorithm agnostic, we can leverage these insights to improve few-shot learning accuracies across many episodic training algorithms.
    We demonstrate the efficacy of our method across popular few-shot learning datasets, algorithms, network architectures, and protocols.
\end{abstract}

\section{Introduction}
\label{sec:introduction}

Large amounts of high-quality data have been the key for the success of deep learning algorithms.
Furthermore, factors such as data augmentation and sampling affect model performance significantly.
Continuously collecting and curating data is a resource (cost, time, storage, etc.) intensive process.
Hence, recently, the machine learning community has been exploring methods for performing transfer-learning from large datasets to unseen tasks with limited data.

A popular genre of these approaches is called meta-learning few-shot approaches, where, in addition to the limited data from the task of interest, a large dataset of disjoint tasks is available for \mbox{(pre-)}training.
These approaches are prevalent in the area of computer vision~\citep{lake2017building} and reinforcement learning~\citep{botvinick2019reinforcement}.
A key component of these methods is the notion of episodic training, which refers to sampling tasks from the larger dataset for training.
By learning to solve these tasks correctly, the model can generalize to new tasks. 

However, sampling for episodic training remains surprisingly understudied despite numerous methods and applications that build on it.
To the best of our knowledge, only a handful of works~\citep{Zhou2020-pn, Sun2019-np, Liu2020-sy} explicitly considered the consequences of sampling episodes.
In comparison, stochastic~\citep{Robbins1951-iy} and mini-batch~\citep{Bertsekas1996-yj} sampling alternatives have been thoroughly analyzed from the perspectives of optimization~\citep{Friedlander2012-lq, Bottou2016-nf}, information theory~\citep{Katharopoulos2018-vh,Csiba2018-bu}, and stochastic processes~\citep{Zhao2014-jf, Zhang2019active}, among many others.
Building a similar understanding of sampling for episodic training will help theoreticians and practitioners develop improved sampling schemes, and is thus of crucial importance to both.

In this paper, we explore many sampling schemes to understand their impact on few-shot methods.
Our work revolves around the following fundamental question: \emph{what is the best way to sample episodes?}
Our focus will be restricted to image classification in the few-shot learning setting -- where ``best'' is taken to mean ``higher transfer accuracy of unseen episodes'' -- and leave analyses and applications to other areas for future work.

Contrary to prior work, our experiments indicate that sampling uniformly with respect to \emph{episode difficulty} yields higher classification accuracy -- a scheme originally proposed to regularize metric learning~\citep{Wu2017-ga}.
To better understand these results, we take a closer look at the properties of episodes and what makes them difficult.
Building on this understanding, we propose a method to approximate different sampling schemes, and demonstrate its efficacy on several standard few-shot learning algorithms and datasets.

Concretely, we make the following contributions:
\begin{itemize}
    \item
        We provide a detailed empirical analysis of episodes and their difficulty.
        When sampled randomly, we show that episode difficulty (approximately) follows a normal distribution and that the difficulty of an episode is largely independent of several modeling choices including the training algorithm, the network architecture, and the training iteration.
    \item 
        Leveraging our analysis, we propose \emph{simple and universally applicable} modifications to the episodic sampling pipeline to approximate \emph{any} sampling scheme.
        We then use this scheme to thoroughly compare episode sampling schemes -- including easy/hard-mining, curriculum learning, and uniform sampling -- and report that sampling uniformly over episode difficulty yields the best results. 
    \item
        Finally, we show that \emph{sampling matters for few-shot classification} as it improves transfer accuracy for a diverse set of popular~\citep{snell2017prototypical,gidaris2018dynamic,finn2017model,raghu2019rapid} and state-of-the-art~\citep{ye2020fewshot} algorithms on standard and cross-domain benchmarks.
\end{itemize}

\section{Preliminaries}
\label{sec:preliminaries}

\subsection{Episodic sampling and training}
\label{subsec:episodic_sampling}

We define episodic sampling as subsampling few-shot tasks (or episodes) from a larger \emph{base} dataset~\citep{chao2020revisiting}.
Assuming the base dataset admits a generative distribution, we sample an episode in two steps\footnote{We use the notation for a probability distribution and its probability density function interchangeably.}.
First, we sample the episode classes $\mathcal{C}_{\tau}$ from class distribution $p(\mathcal{C}_{\tau})$; second, we sample the episode's data from data distribution $p(x, y \mid \mathcal{C}_{\tau})$ conditioned on $\mathcal{C}_{\tau}$.
This gives rise to the following log-likelihood for a model $l_\theta$ parameterized by $\theta$:
\begin{align}
    \label{eq:likelihood}
    \mathcal{L}(\theta) 
    &= \Exp[\tau \sim q(\cdot)]{\log l_\theta(\tau)},
\end{align}
where $q(\tau)$ is the episode distribution induced by first sampling classes, then data.
In practice, this expectation is approximated by sampling a batch of episodes $\mathcal{B}$ each with their set $\tau_Q$ of \emph{query} samples.
To enable transfer to unseen classes, it is also common to include a small set $\tau_S$ of \emph{support} samples to provide statistics about $\tau$.
This results in the following Monte-Carlo estimator:
\begin{equation}
    \label{eq:montecarlo}
    \mathcal{L}(\theta) \approx \frac1{\vert \mathcal{B} \vert} \sum_{\tau \in \mathcal{B}} \frac1{\vert \tau_Q \vert} \sum_{(x, y) \in \tau_Q} \log l_\theta (y \mid x, \tau_S),
\end{equation}
where the data in $\tau_Q$ and $\tau_S$ are both distributed according to $p(x, y \mid \mathcal{C}_\tau)$.
In few-shot classification, the $n$-way $k$-shot setting corresponds to sampling $n = \vert \mathcal{C_\tau} \vert$ classes, each with $k = \frac{\vert \tau_S \vert}{n}$ support data points.
The implicit assumption in the vast majority of few-shot methods is that both classes and data are sampled with \emph{uniform} probability -- \emph{but are there better alternatives?}
We carefully examine these underlying assumptions in the forthcoming sections.

\subsection{Few-shot algorithms}

We briefly present a few representative episodic learning algorithms.
A more comprehensive treatment of few-shot algorithms is presented in~\citet{Wang2020-fj} and~\citet{Hospedales2020-ln}.
A core question in few-shot learning lies in evaluating (and maximizing) the model likelihood $l_\theta$.
These algorithms can be divided in two major families: gradient-based methods, which adapt the model's parameters to the episode; and metric-based methods, which compute similarities between support and query samples in a learned embedding space.

Gradient-based few-shot methods are best illustrated through Model-Agnostic Meta-Learning~\citep{finn2017model} (MAML).
The intuition behind MAML is to learn a set of initial parameters which can quickly specialize to the task at hand.
To that end, MAML computes $l_\theta$ by adapting the model parameters $\theta$ via one (or more) steps of gradient ascent and then computes the likelihood using the adapted parameters $\theta'$.
Concretely, we first compute the likelihood $p_\theta(y \mid x)$ using the support set $\tau_S$, adapt the model parameters, and then evaluate the likelihood:
\begin{align*}
    & l_\theta(y \mid x, \tau_S) = p_{\theta'}(y \mid x) \quad \text{s.t.} \quad \theta' = \theta + \alpha \nabla_\theta \sum_{(x, y) \in \tau_S} \log p_\theta(y \mid x),
\end{align*}
where $\alpha > 0$ is known as the adaptation learning rate.
A major drawback of training with MAML lies in back-propagating through the adaptation phase, which requires higher-order gradients.
To alleviate this computational burden, Almost No Inner Loop~\citep{raghu2019rapid} (ANIL) proposes to only adapt the last classification layer of the model architecture while tying the rest of the layers across episodes.
They empirically demonstrate little classification accuracy drop while accelerating training times four-fold.

Akin to ANIL, metric-based methods also share most of their parameters across tasks; however, their aim is to learn a metric space where classes naturally cluster.
To that end, metric-based algorithms learn a feature extractor $\phi_\theta$ parameterized by $\theta$ and classify according to a non-parametric rule.
A representative of this family is Prototypical Network~\citep{snell2017prototypical} (ProtoNet), which classifies query points according to their distance to \emph{class prototypes} -- the average embedding of a class in the support set:
\begin{align*}
    l_\theta(y \mid x, \tau_S)
    = \frac{\exp \left( -d(\phi_\theta(x), \phi_\theta^y) \right)}{\sum_{y' \in \mathcal{C}_{\tau}} \exp \left( - d(\phi_\theta(x), \phi_\theta^{y'}) \right)} \quad \text{s.t.} \quad \phi_\theta^c = \frac1{k} \sum_{\substack{(x, y) \in \tau_S \\ y = c}} \phi_\theta(x),
\end{align*}
where $d(\cdot, \cdot)$ is a distance function such as the Euclidean distance or the negative cosine similarity, and $\phi_\theta^c$ is the class prototype for class $c$.
Other classification rules include support vector clustering~\citep{lee2019meta}, neighborhood component analysis~\citep{Laenen2020-oo}, and the earth-mover distance~\citep{zhang2020deepemd}.

\subsection{Episode difficulty}
\label{subsec:episode_difficulty}

Given an episode $\tau$ and likelihood function $l_\theta$, we define \emph{episode difficulty} to be the negative log-likelihood incurred on that episode:
\begin{equation*}
    \Omega_{l_\theta}(\tau) = - \log l_\theta (\tau),
\end{equation*}
which is a surrogate for how hard it is to classify the samples in $\tau_Q$ correctly, given $l_\theta$ and $\tau_S$.
By definition, this choice of episode difficulty is tied to the choice of the likelihood function $l_\theta$.

\citet{dhillon2019baseline} use a similar surrogate as a means to systematically report few-shot performances.
We use this definition because it is equivalent to the loss associated with the likelihood function $l_{\theta}$ on episode $\tau$, which is readily available at training time.

\section{Methodology}
\label{sec:methodology}

In this section, we describe the core assumptions and methodology used in our study of sampling methods for episodic training.
Our proposed method builds on importance sampling~\citep{glasserman2004monte} (IS) which we found compelling for three reasons: (i) IS is \emph{well understood} and solidly grounded from a theoretical standpoint, (ii) IS is \emph{universally applicable} thus compatible with all episodic training algorithms, and (iii) IS is \emph{simple to implement} with little requirement for hyper-parameter tuning.

\emph{Why should we care about episodic sampling?}
A back-of-the-envelope calculation\footnote{For a base dataset with $N$ classes and $K$ input-output pairs per class, there are a total of $\binom{N}{n} \binom{K}{k}^{n}$ possible episodes that can be created when sampling $k$ pairs each from $n$ classes.}
suggests that there are on the order of $10^{162}$ different training episodes for the smallest-scale experiments in \cref{sec:experiments}.
Since iterating through each of them is infeasible, we ought to express some preference over which episodes to sample.
In the following, we describe a method that allows us to specify this preference.

\subsection{Importance sampling for episodic training}
\label{subsec:is}

\begin{algorithm}[t]
    \caption{Episodic training with Importance Sampling}
    \label{alg:is-ess}
    \begin{algorithmic}
        \STATE {\bfseries Input:} target ($p$) and proposal ($q$) distributions, likelihood function $l_{\theta}$, optimizer OPT.
        \STATE Randomly initialize model parameters $\theta$.
        \REPEAT
        \STATE Sample a mini-batch $\mathcal{B}$ of episodes from $q(\tau)$.
        \FOR{each episode $\tau$ in mini-batch $\mathcal{B}$}
        \STATE Compute episode likelihood: $l_\theta(\tau)$.
        \STATE Compute importance weight: $w(\tau) = \frac{p(\tau)}{q(\tau)}$.
        \ENDFOR
        \STATE Aggregate: $\mathcal{L}(\theta) \gets \sum_{\tau \in \mathcal{B}} w(\tau) \log l_\theta(\tau)$.
        \STATE Compute effective sample size $\ess(\mathcal{B})$.
        \STATE Update model parameters: $\theta \gets \text{OPT}(\frac{\mathcal{L}(\theta)}{\ess(\mathcal{B})} )$.
        \UNTIL{parameters $\theta$ have converged.}
    \end{algorithmic}
\end{algorithm}

Let us assume that the sampling scheme described in~\cref{subsec:episodic_sampling} induces a distribution $q(\tau)$ over episodes.
We call it the \emph{proposal distribution}, and assume knowledge of its density function.
We wish to estimate the expectation in~\cref{eq:likelihood} when sampling episodes according to a \emph{target distribution} $p(\tau)$ of our choice, rather than $q(\tau)$.
To that end, we can use an importance sampling estimator which simply re-weights the observed values for a given episode $\tau$ by $w(\tau) = \frac{p(\tau)}{q(\tau)}$, the ratio of the target and proposal distributions:
\begin{align*}
    \Exp[\tau \sim p(\cdot)]{\log l_\theta(\tau)}
        &= \Exp[\tau \sim q(\cdot)]{w(\tau) \log l_\theta(\tau)}.
\end{align*}
The importance sampling identity holds whenever $q(\tau)$ has non-zero density over the support of $p(\tau)$, and effectively allows us to sample from \emph{any} target distribution $p(\tau)$.

A practical issue of the IS estimator arises when some values of $w(\tau)$ become much larger than others; in that case, the likelihoods $l_\theta(\tau)$ associated with mini-batches containing heavier weights dominate the others, leading to disparities.
To account for this effect, we can replace the mini-batch average in the Monte-Carlo estimate of~\cref{eq:montecarlo} by the \emph{effective sample size} $\ess(\mathcal{B})$~\citep{kong1992note,liu1996metropolized}:
\begin{align}
    \label{eq:mc-ess}
    \Exp[\tau \sim p(\cdot)]{\log l_\theta(\tau)} \approx \frac1{\ess(\mathcal{B})}\sum_{\tau \in \mathcal{B}} w(\tau) \log l_\theta(\tau) \quad \text{s.t.} \quad \ess(\mathcal{B}) = \frac{(\sum_{\tau \in \mathcal{B}} w(\tau))^{2}}{\sum_{\tau \in \mathcal{B}} w(\tau)^{2}},
\end{align}
where $\mathcal{B}$ denotes a mini-batch of episodes sampled according to $q(\tau)$.
Note that when $w(\tau)$ is constant, we recover the standard mini-batch average setting as $\ess(\mathcal{B}) = \vert\mathcal{B}\vert$.
Empirically, we observed that normalizing with the effective sample size avoided instabilities.
This method is summarized in~\cref{alg:is-ess}.

\subsection{Modeling the proposal distribution}
\label{subsec:proposal_distribution}

A priori, we do not have access to the proposal distribution $q(\tau)$ (nor its density) and thus need to estimate it empirically.
Our main assumption is that sampling episodes from $q(\tau)$ induces a normal distribution over episode difficulty.
With this assumption, we model the proposal distribution by this induced distribution, therefore replacing $q(\tau)$ with $\mathcal{N}(\Omega_{l_\theta}(\tau) \mid \mu, \sigma^2)$ where $\mu, \sigma^2$ are the mean and variance parameters.
As we will see in~\cref{subsec:episode_hardness}, this normality assumption is experimentally supported on various datasets, algorithms, and architectures.

We consider two settings for the estimation of $\mu$ and $\sigma^2$: offline and online.
The \emph{offline} setting consists of sampling $1,000$ training episodes before training, and computing $\mu, \sigma^{2}$ using a model pre-trained on the same base dataset. Though this setting seems  unrealistic, \ie  having access to a pre-trained model, several meta-learning few-shot methods start with a pre-trained model which they further build upon.
Hence, for such methods there is no overhead.
For the \emph{online} setting, we estimate the parameters on-the-fly using the model currently being trained.
This is justified by the analysis in~\cref{subsec:episode_hardness} which shows that episode difficulty transfers across model parameters during training.
We update our estimates of $\mu, \sigma^{2}$ with an exponential moving average:
\begin{align*}
    \mu \gets \lambda \mu + (1 - \lambda) \Omega_{l_\theta}(\tau) \quad \text{and} \quad \sigma^{2} \gets \lambda \sigma^2 + (1 - \lambda) (\Omega_{l_\theta}(\tau) - \mu)^2,
\end{align*}
where $\lambda \in [0, 1]$ controls the adjustment rate of the estimates, and the initial values of $\mu, \sigma^2$ are computed in a warm-up phase lasting $100$ iterations.
Keeping $\lambda = 0.9$ worked well for all our experiments (\cref{sec:experiments}).
We opted for this simple implementation as more sophisticated approaches like~\citet{west1979updating} yielded little to no benefit.

\subsection{Modeling the target distribution}
\label{subsec:target_distribution}

Similar to the proposal distribution, we model the target distribution by its induced distribution over episode difficulty.
Our experiments compare four different approaches, all of which share parameters $\mu, \sigma^2$ with the normal model of the proposal distribution.
For numerical stability, we truncate the support of all distributions to $[\mu - 2.58\sigma, \mu + 2.58\sigma]$, which gives approximately $99\%$ coverage for the normal distribution centered around $\mu$.

The first approach (\hard) takes inspiration from hard negative mining~\citep{shrivastava2016training}, where we wish to sample only more challenging episodes.
The second approach (\easy) takes a similar view but instead only samples easier episodes.
We can model both distributions as follows:
\begin{align*}
    & \mathcal{U}(\Omega_{l_\theta}(\tau) \mid \mu, \mu + 2.58\sigma) \tag{\hard}
    \\
    & \qquad \qquad \text{and}
    \\
    & \mathcal{U}(\Omega_{l_\theta}(\tau) \mid \mu - 2.58\sigma, \mu) \tag{\easy}
\end{align*}
where $\mathcal{U}$ denotes the uniform distribution.
The third (\curriculum) is motivated by curriculum learning~\citep{Bengio2009-og}, which slowly increases the likelihood of sampling more difficult episodes:
\begin{equation*}
    \mathcal{N}(\Omega_{l_\theta}(\tau) \mid \mu_t, \sigma^2)
    \tag{\curriculum}
\end{equation*}
where $\mu_t$ is linearly interpolated from $\mu - 2.58\sigma$ to $\mu + 2.58\sigma$ as training progresses.
Finally, our fourth approach, \uniform, resembles distance weighted sampling~\citep{Wu2017-ga} and consists of sampling uniformly over episode difficulty:
\begin{equation*}
    \mathcal{U}(\Omega_{l_\theta}(\tau) \mid \mu - 2.58\sigma, \mu + 2.58\sigma).  \tag{\uniform}
\end{equation*}
Intuitively, \uniform can be understood as a uniform prior over unseen test episodes, with the intention of performing well across the entire difficulty spectrum.
This acts as a regularizer, forcing the model to be equally discriminative for both easy and hard episodes.

\section{Related Works}

This paper studies task sampling in the context of few-shot~\citep{Miller2000-hy, Fei-Fei2006-od} and meta-learning~\citep{Schmidhuber1987-cx, Thrun1998-lt}.

\paragraph{Few-shot learning.}

This setting has received a lot of attention over recent years~\citep{vinyals2016matching,ravi2016optimization, Santoro2016-br, Garnelo2018-cx}.
Broadly speaking, state-of-the-art methods can be categorized in two major families: metric-based and gradient-based.

Metric-based methods learn a shared feature extractor which is used to compute the distance between samples in embedding space~\citep{snell2017prototypical, Bertinetto2018-gn, Ravichandran2019-jr, Laenen2020-oo}.
The choice of metric mostly differentiates one method from another; for example, popular choices include Euclidean distance~\citep{snell2017prototypical}, negative cosine similarity~\citep{gidaris2018dynamic}, support vector machines~\citep{lee2019meta}, set-to-set functions~\citep{ye2020fewshot}, or the earth-mover distance~\cite{zhang2020deepemd}.

Gradient-based algorithms such as MAML~\citep{finn2017model}, propose an objective to learn a network initialization that can quickly adapt to new tasks.
Due to its minimal assumptions, MAML has been extended to probabilistic formulations~\citep{Grant2018-lh, Yoon2018-om} to incorporate learned optimizers -- implicit~\citep{Flennerhag2019-fw} or explicit~\citep{Park2019-or} -- and simplified to avoid expensive second-order computations~\citep{Nichol2018-wm, Rajeswaran2019-hx}.
In that line of work, ANIL~\citep{raghu2019rapid} claims to match MAML’s performance when adapting only the last classification layer -- thus greatly reducing the computational burden and bringing gradient and metric-based methods closer together.

\paragraph{Sampling strategies.}

Sampling strategies have  been studied for different training regimes.
\citet{Wu2017-ga} demonstrate that ``sampling matters'' in the context of metric learning.
They propose to sample a triplet with probability proportional to the distance of its positive and negative samples, and observe stabilized training and improved accuracy.
This observation was echoed by~\citet{Katharopoulos2018-vh} when sampling mini-batches: carefully choosing the constituent samples of a mini-batch improves the convergence rate and asymptotic performance.
Like ours, their method builds on importance sampling~\citep{Smith1997-cs, Doucet2001-zu, Johnson2018-ki} but whereas they compute importance weights using the magnitude of the model's gradients, we use the episode's difficulty.
Similar insights were also observed in reinforcement learning, where~\citet{Schaul2015-rv} suggests a scheme to sample transitions according to the temporal difference error.

Closer to our work,~\citet{Sun2019-np} present a hard-mining scheme where the most challenging classes across episodes are pooled together and used to create new episodes.
Observing that the difficulty of a class is intrinsically linked to the other classes in the episode,~\citet{Liu2020-sy} propose a mechanism to track the difficulty across every class pair.
They use this mechanism to build a curriculum~\citep{Bengio2009-og, Wu2020-xe} of increasingly difficult episodes.
In contrast to these two approaches, our proposed method makes use of importance sampling to mimic the target distribution rather than sampling from it directly.
This helps achieve fast and efficient sampling without any preprocessing requirements.

\section{Experiments}
\label{sec:experiments}

We first validate the assumptions underlying our proposed IS estimator and shed light on the properties of episode difficulty.
Then, we answer the question we pose in the introduction, namely: \emph{what is the best way to sample episodes?}
Finally, we ask if better sampling improves few-shot classification.

\subsection{Experimental setup} \label{subsec:exp_setup}

We review the standardized few-shot benchmarks and provide a detailed description in~\cref{app:setup}.

\paragraph{Datasets.}
We use two standardized image classification datasets, \mimg~\citep{vinyals2016matching} and \timg~\citep{ren2018meta}, both subsets of ImageNet~\citep{deng2009imagenet}.
\mimg consists of $64$ classes for training, $16$ for validation, and $20$ for testing; we use the class splits introduced by~\citet{ravi2016optimization}.
\timg contains $608$ classes split into $351$, $97$, and $160$ for training, validation, and testing, respectively.

\paragraph{Network architectures.}
We train two model architectures.
A 4-layer convolution network \cnn~\citet{vinyals2016matching} with $64$ channels per layer.
And \resnet, a 12-layer deep residual network~\citep{he2016deep} introduced by~\citet{oreshkin2018tadam}.
Both architectures use batch normalization~\citep{ioffe2015batch} after every convolutional layer and ReLU as the non-linearity.

\paragraph{Training algorithms.}
For the metric-based family, we use ProtoNet with Euclidean~\citep{snell2017prototypical} and scaled negative cosine similarity measures~\citep{gidaris2018dynamic}.
Additionally, we use MAML~\citep{finn2017model} and ANIL~\citep{raghu2019rapid} as representative gradient-based algorithms.

\paragraph{Hyper-parameters.}
We tune hyper-parameters for each algorithm and dataset to work well across different few-shot settings and network architectures.
Additionally, we keep the hyper-parameters the same across all different sampling methods for a fair comparison.
We train for $20$k iterations with a mini-batch of size $16$ and $32$ for \mimg and \timg respectively, and validate every $1$k iterations on $1$k episodes.
The best performing model is finally evaluated on $1$k test episodes.

\subsection{Understanding episode difficulty}
\label{subsec:episode_hardness}

All the models in this subsection are trained using baseline sampling as described in~\cref{subsec:episodic_sampling}, \ie, episodic training without importance sampling.

\begin{figure}[t]
    \centering
    \begin{subfigure}
        \centering
        \includegraphics[height=.26\linewidth]{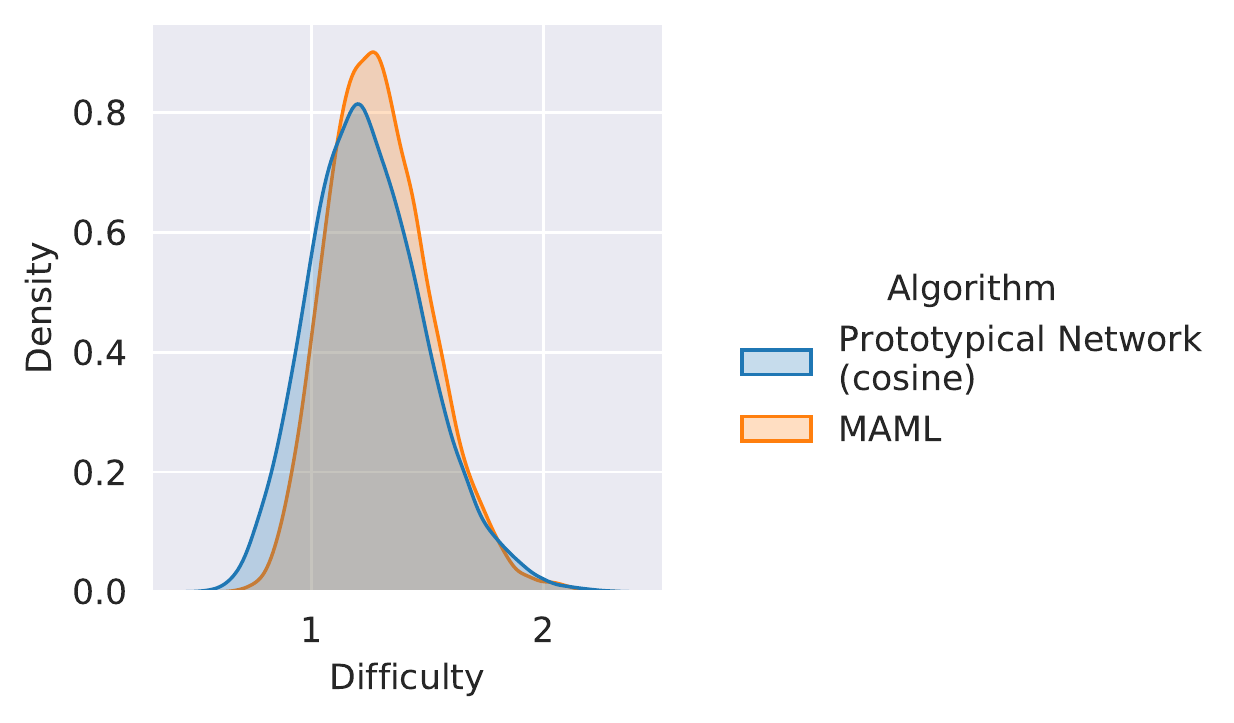}
    \end{subfigure}
    \begin{subfigure}
        \centering
        \includegraphics[height=.26\linewidth]{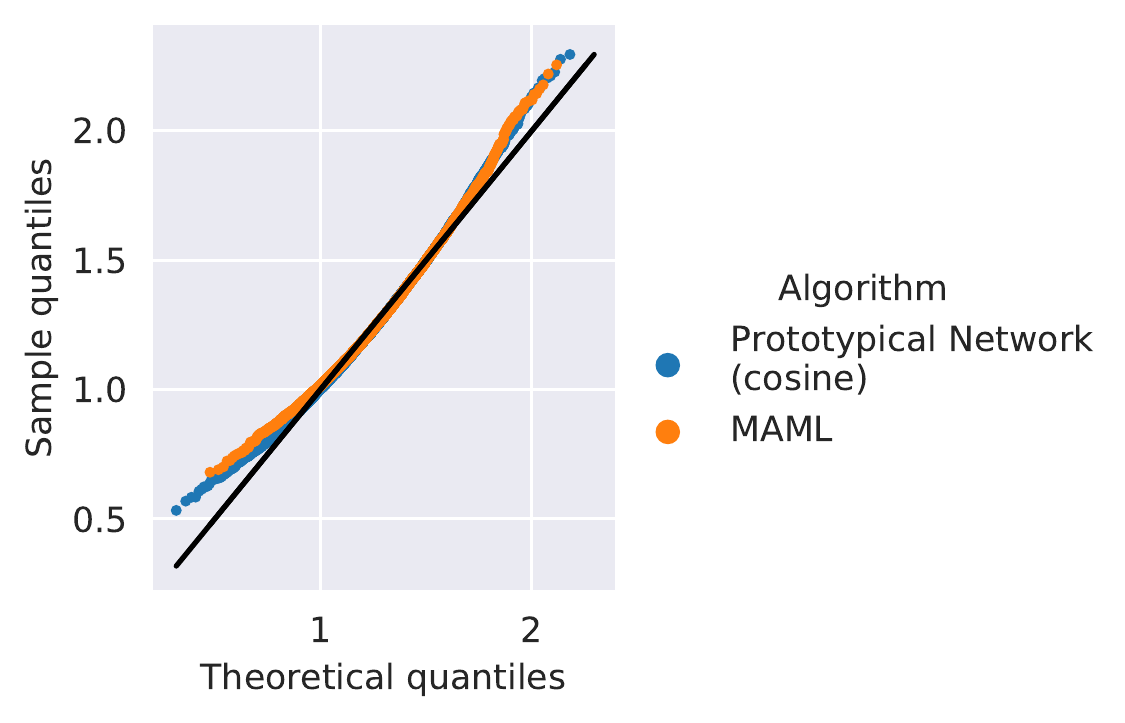}
    \end{subfigure}
    \caption{
        \small
        \textbf{Episode difficulty is approximately normally distributed.}
        Density (left) and Q-Q (right) plots of the episode difficulty computed by \cnn's on \mimg (\osfw), trained using ProtoNets (cosine) and MAML (depicted in the legends).
        The values are computed over $10$k test episodes.
        The density plots follow a bell curve, with the density peak in the middle, which quickly drops-off on either side of the peak.
        The Q-Q plots are close to the identity line (in black).
        The closer the curve is to the identity line, the closer the distribution is to a normal.
        Both suggest that the episode difficulty distribution can be normally approximated.
    }
    \label{fig:hardness_normal}
    \vspace{-1em}
\end{figure}

\subsubsection{Episode difficulty is approximately normally distributed}
\label{subsubsec:episode_hardness_normal}

We begin our analysis by verifying that the distribution over episode difficulty induced by $q(\tau)$ is approximately normal.
In~\cref{fig:hardness_normal}, we use the difficulty of $10$k test episodes sampled with $q(\tau)$.
The difficulties are computed using \cnn trained with ProtoNet and MAML on \mimg for \osfw classification.
The episode difficulty density plots follow a bell curve, which are naturally modeled with a normal distribution.
The Q-Q plots, typically used to assess normality, suggest the same -- the closer the curve is to the identity line, the closer the distribution is to a normal.

Finally, we compute the Shapiro-Wilk test for normality~\citep{shapiro1965analysis}, which tests for the null hypothesis that the data is drawn from a normal distribution.
Since the p-value for this test is sensitive to the sample size\footnote{For a large sample size, the p-values are not reliable as they may detect trivial departures from normality.}, we subsample $50$ values $100$ times and average rejection rates over these subsets.
With $\alpha = 0.05$, the null hypothesis is rejected $14\%$ and $17\%$ of the time for \mimg and \timg respectively, thus suggesting that episode difficulty can be reliably approximated with a normal distribution.

\subsubsection{Independence from modeling choices}
\label{subsubsec:episode_hardness_model}

By definition, the notion of episode difficulty is tightly coupled to the model likelihood $l_{\theta}$ (\cref{subsec:episode_difficulty}), and hence to the modeling variables such as learning algorithm, network architecture, and model parameters.
We check if episode difficulty transfers across different choices for these variables.

\begin{wrapfigure}{l}{0.6\linewidth}
    \vspace{-1.7em}
    \centering
    \includegraphics[width=1.0\linewidth]{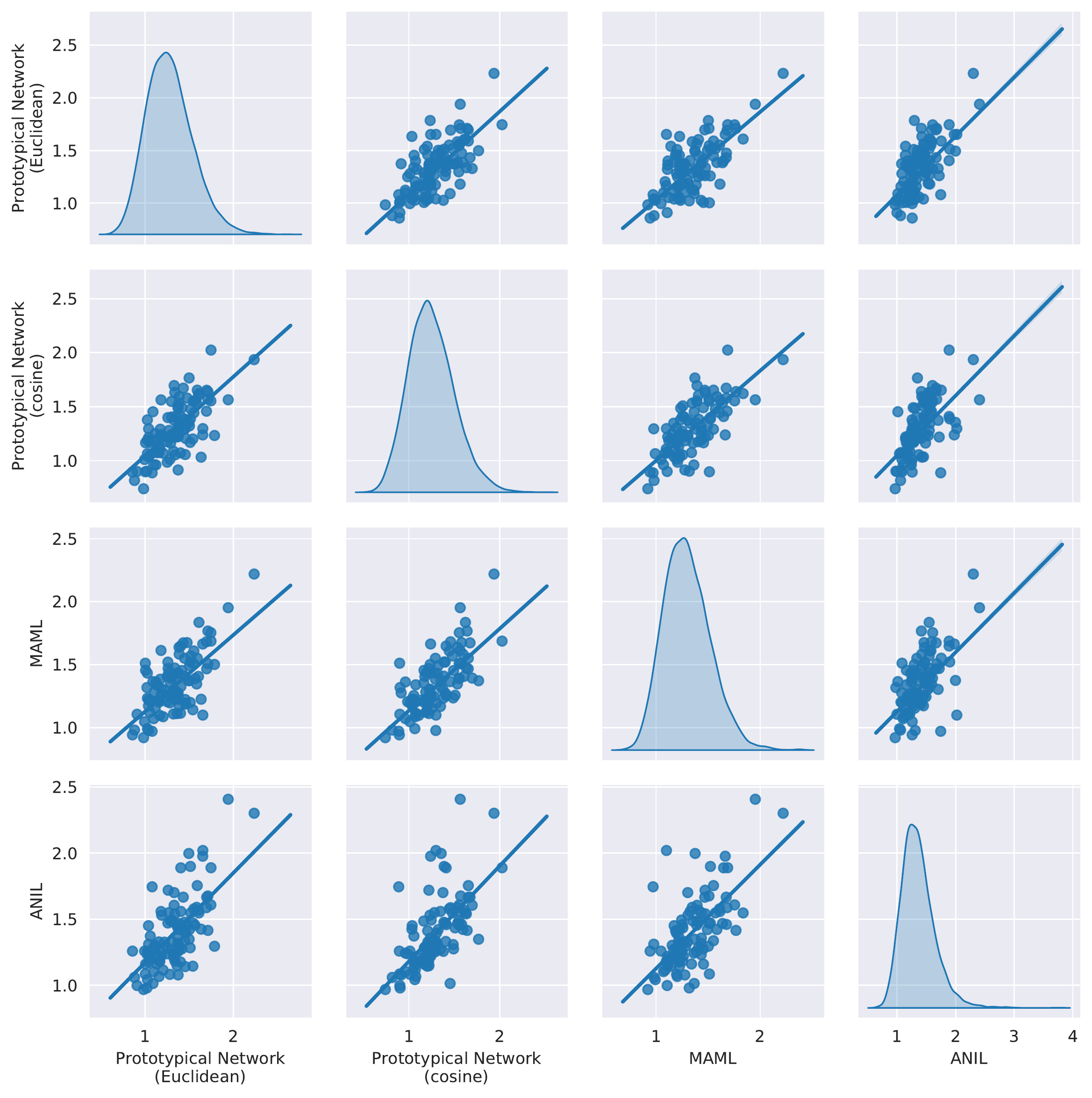}
    \caption{
        \small
        \textbf{Episode difficulty transfers across training algorithms.}
        Scatter plots (with regression lines) of the episode difficulty computed on $1$k \mimg test episodes (\osfw) by \cnn's trained using different algorithms.
        The positive correlation suggests that an episode that is difficult for one training algorithm will be difficult for another.
    }
    \label{fig:hardness_algorithm}
    \vspace{-3em}
\end{wrapfigure}

We are concerned with the \emph{relative ranking} of the episode difficulty and not the actual values.
To this end, we will use the Spearman rank-order correlation coefficient, a non-parametric measure of the monotonicity of the relationship between two sets of values.
This value lies within $[-1; 1]$, with $0$ implying no correlation, and $+1$ and $-1$ implying exact positive and negative correlations, respectively.

\paragraph{Training algorithm.}

We first check the dependence on the training algorithm.
We use all four algorithms to train \cnn's for \osfw classification on \mimg, then compute episode difficulty over $10$k test episodes.
The Spearman rank-order correlation coefficients for the difficulty values computed with respect to all possible pairs of training algorithms are $>0.65$.
This positive correlation is illustrated in~\cref{fig:hardness_algorithm} and suggests that an episode that is difficult for one training algorithm is very likely to be difficult for another.

\paragraph{Network architecture.}

\begin{figure}[t]
\noindent
\begin{minipage}{.465\textwidth}
    \centering
    \vspace{-2.5em}
    \includegraphics[width=1.0\linewidth]{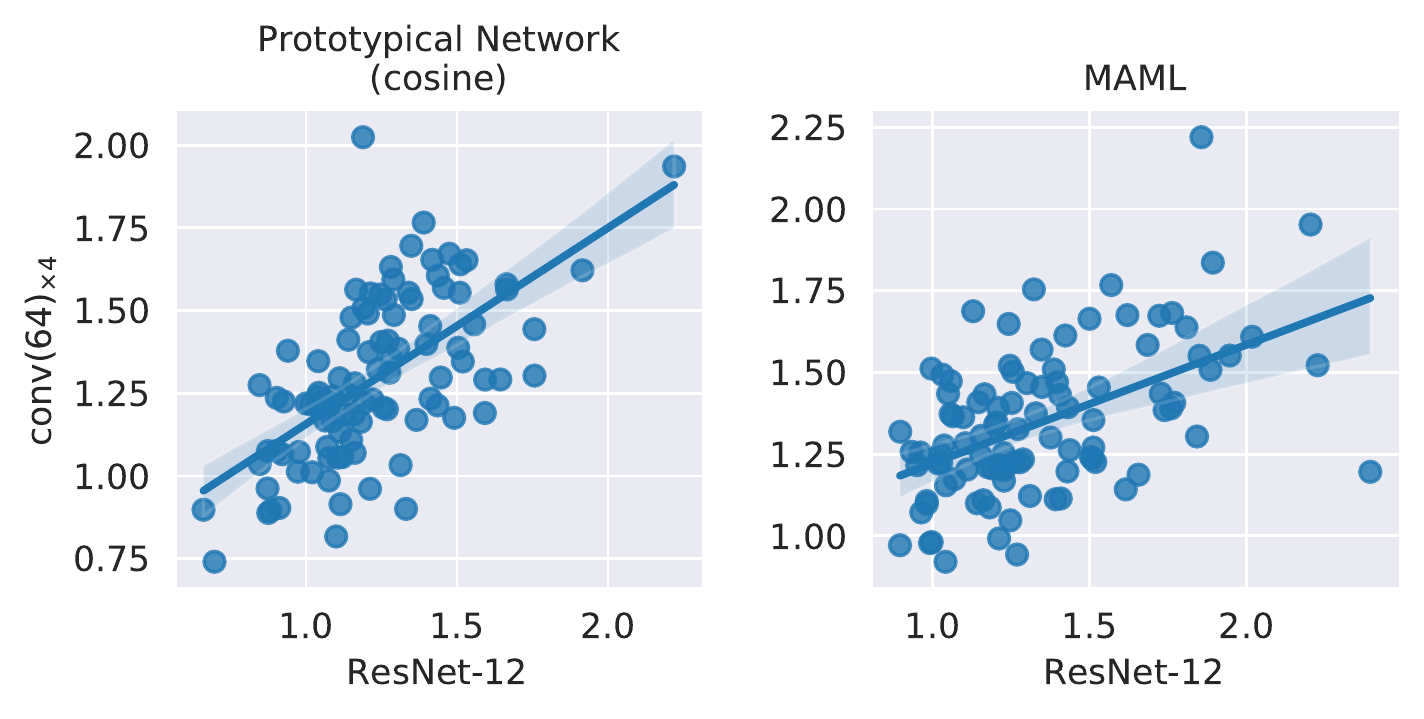}
    \captionof{figure}{
        \small
        \textbf{Episode difficulty transfers across network architectures.}
        Scatter-plots (with regression lines) of the episode difficulty computed by \cnn and \resnet's trained using different algorithms.
        This is computed for $10$k \osfw test episodes from \mimg.
        We observe a strong positive correlation between the computed values for both network architectures.
    }
    \label{fig:hardness_model}
\end{minipage}
\hspace{.005\textwidth}
\begin{minipage}{.51\textwidth}
    \centering
    \includegraphics[width=1.0\linewidth]{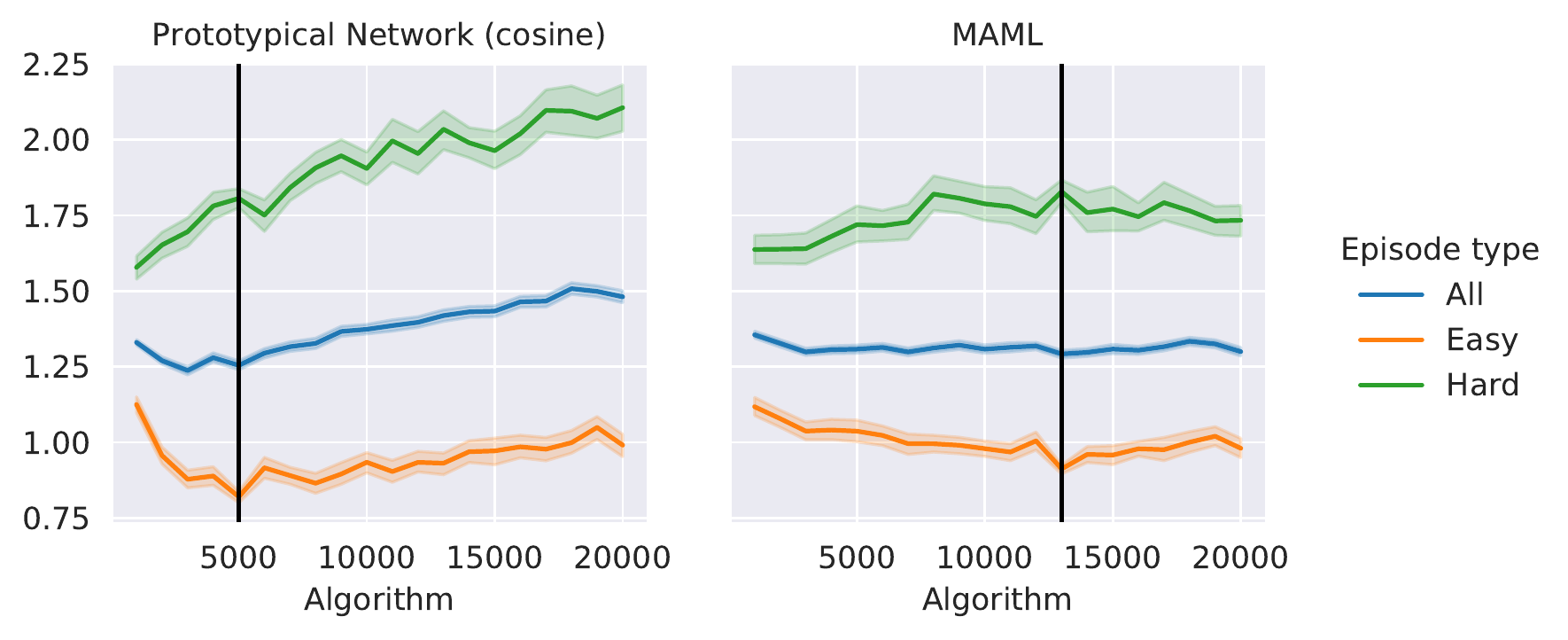}
    \vspace{-0.5em}
    \captionof{figure}{
        \small
        \textbf{Episode difficulty is transferred across model parameters during training.}
        We select the $50$ easiest and hardest episodes and track their difficulty during training.
        This is done for \cnn's trained on \mimg (\osfw) with different algorithms.
        The average difficulty of the episodes decreases over time, until convergence (vertical line), after which the model overfits.
        Additionally, easier episodes remain easy while harder episodes remain hard, indicating that episode difficulty transfers from one set of parameters to the next.
    }
    \label{fig:hardness_time}
\end{minipage}
\vspace{-1.0em}
\end{figure}

Next we analyze the dependence on the network architecture.
We trained \cnn and \resnet's using all training algorithms for \mimg \osfw classification.
We compute the episode difficulties for $10$k test episodes and compute their Spearman rank-order correlation coefficients across the two architectures, for a given algorithm.
The correlation coefficients are $0.57$ for ProtoNet (Euclidean), $0.72$ for ProtoNet (cosine), $0.58$ for MAML, and $0.49$ for ANIL.
\cref{fig:hardness_model} illustrates this positive correlation, suggesting that episode difficulty is transferred across network architectures with high probability.

\paragraph{Model parameters during training.}

Lastly, we study the dependence on model parameters during training.
We select the $50$ \emph{easiest} and $50$ \emph{hardest} episodes, \ie, episodes with the lowest and highest difficulties respectively, from $1$k test episodes.
We track the episode difficulty for all episodes over the training phase and visualize the trend in~\cref{fig:hardness_time} for \cnn's trained using different training algorithms on \mimg (\osfw).
Throughout training, easy episodes remain easy and hard episodes remain hard, hence suggesting that episode difficulty transfers across different model parameters during training.
Since the episode difficulty does not change drastically during the training process, we can estimate it with a running average over the training iterations.
This justifies the \emph{online} modeling of the proposal distribution in~\cref{subsec:proposal_distribution}.

\begin{table*}[ht]
\centering
\caption{
    \small
    \textbf{Few-shot accuracies on benchmark datasets for 5-way few-shot episodes in the offline setting.}
    Mean accuracy and $95\%$ confidence interval computed over $1$k test episodes.
    The first row in every scenario denotes baseline sampling.
    Best results for a fixed scenario are shown in bold and the $\dagger$ indicates matching or improving over baseline sampling.
}
\label{tab:offline_proto}
\setlength{\tabcolsep}{3pt}
{
\small
\begin{tabular}{l cc cc cc}
\toprule

& \multicolumn{4}{c}{\mimg} & \multicolumn{2}{c}{\timg} \\
\cmidrule(lr){2-5} \cmidrule(lr){6-7}

& \multicolumn{2}{c}{\cnn}  & \multicolumn{2}{c}{\resnet} & \multicolumn{2}{c}{\resnet} \\
\cmidrule(lr){2-3} \cmidrule(lr){4-5} \cmidrule(lr){6-7}

& 1-shot (\%) & 5-shot (\%) & 1-shot (\%) & 5-shot (\%) & 1-shot (\%) & 5-shot (\%) \\

\midrule

ProtoNet (cosine) &
\textbf{50.03$\pm$0.61} & 61.56$\pm$0.53 &
52.85$\pm$0.64 & 62.11$\pm$0.52 &
\textbf{60.01$\pm$0.73} & 72.75$\pm$0.59
\\

+ \easy &
\textbf{49.60$\pm$0.61}$^{\dagger}$ & 65.17$\pm$0.53$^{\dagger}$ &
53.35$\pm$0.63$^{\dagger}$ & 63.55$\pm$0.53$^{\dagger}$ &
\textbf{60.03$\pm$0.75}$^{\dagger}$ & 74.65$\pm$0.57$^{\dagger}$
\\

+ \hard &
49.01$\pm$0.60 & \textbf{66.45$\pm$0.50}$^{\dagger}$ &
52.65$\pm$0.63$^{\dagger}$ & 70.15$\pm$0.51$^{\dagger}$ &
55.44$\pm$0.72 & 75.97$\pm$0.55$^{\dagger}$
\\

+ \curriculum &
49.38$\pm$0.61 & 64.12$\pm$0.53$^{\dagger}$ &
53.21$\pm$0.65$^{\dagger}$ & 65.89$\pm$0.52$^{\dagger}$ &
\textbf{60.37$\pm$0.76}$^{\dagger}$ & 75.32$\pm$0.58$^{\dagger}$
\\

+ \uniform &
\textbf{50.07$\pm$0.59}$^{\dagger}$ & \textbf{66.33$\pm$0.52}$^{\dagger}$ &
\textbf{54.27$\pm$0.65}$^{\dagger}$ & \textbf{70.85$\pm$0.51}$^{\dagger}$ &
\textbf{60.27$\pm$0.75}$^{\dagger}$ & \textbf{78.36$\pm$0.54}$^{\dagger}$
\\

\bottomrule
\end{tabular}
}
\vspace{-1em}
\end{table*}

\subsection{Comparing episode sampling methods}
\label{subsec:offline}

We compare different methods for episode sampling.
To ensure fair comparisons, we use the offline formulation (\cref{subsec:proposal_distribution}) so that all sampling methods share the same pre-trained network (the network trained using baseline sampling) when computing proposal likelihoods.
We compute results over $2$ datasets, $2$ network architectures, $4$ algorithms and $2$ few-shot protocols, totaling in $24$ scenarios.
\cref{tab:offline_proto} presents results on ProtoNet (cosine), while the rest are in~\cref{app:offline}.

We observe that, although not strictly dominant, \uniform tends to outperform other methods as it is within the statistical confidence of the best method in $19/24$ scenarios.
For the $5/24$ scenarios where \uniform underperforms, it closely trails behind the best methods.
Compared to baseline sampling, the average degradation of \uniform is $-0.83\%$ and at most $-1.44\%$ (ignoring the standard deviations) in $4/24$ scenarios.
Conversely, \uniform boosts accuracy by as much as $8.74\%$ and on average by $3.86\%$ (ignoring the standard deviations) in $13/24$ scenarios.
We attribute this overall good performance to the fact that uniform sampling puts a uniform distribution prior over the (unseen) test episodes, with the intention of performing well across the entire difficulty spectrum.
This acts as a regularizer, forcing the model to be equally discriminative for easy and hard episodes.
If we knew the test episode distribution, upweighting episodes that are most likely under that distribution will improve transfer accuracy~\citep{fallah2021generalization}.
However, this uninformative prior is the safest choice without additional information about the test episodes.

Second best is baseline sampling as it is statistically competitive on $10/24$ scenarios, while \easy, \hard, and \curriculum only appear among the better methods in $4$, $4$, and $9$ scenarios, respectively.

\begin{table*}[t]
\centering
\caption{
    \small
    \textbf{Few-shot accuracies on benchmark datasets for 5-way few-shot episodes in the offline and online settings.}
    The mean accuracy and the $95\%$ confidence interval are reported for evaluation done over $1$k test episodes.
    The first row in every scenario denotes baseline sampling.
    Best results for a fixed scenario are shown in bold.
    Results where a sampling technique is better than or comparable to baseline sampling are denoted by $\dagger$.
    \uniform (Online) retains most of the performance of the offline formulation while being significantly easier to implement (online is competitive in $15/24$ scenarios vs $16/24$ for offline).
    }
\label{tab:online_full}
\setlength{\tabcolsep}{2pt}
{
\small
\begin{tabular}{l cc cc cc}
\toprule

& \multicolumn{4}{c}{\mimg} & \multicolumn{2}{c}{\timg} \\
\cmidrule(lr){2-5} \cmidrule(lr){6-7}

& \multicolumn{2}{c}{\cnn}  & \multicolumn{2}{c}{\resnet} & \multicolumn{2}{c}{\resnet} \\
\cmidrule(lr){2-3} \cmidrule(lr){4-5} \cmidrule(lr){6-7}

& 1-shot (\%) & 5-shot (\%) & 1-shot (\%) & 5-shot (\%) & 1-shot (\%) & 5-shot (\%) \\

\midrule

ProtoNet (Euclidean) &
\textbf{49.06$\pm$0.60} & 65.28$\pm$0.52 &
49.67$\pm$0.64 & 67.45$\pm$0.51 &
\textbf{59.10$\pm$0.73} & 76.95$\pm$0.56
\\

+ \uniform (Offline) &
48.19$\pm$0.62 & 66.73$\pm$0.52$^{\dagger}$ &
\textbf{53.94$\pm$0.63}$^{\dagger}$ & \textbf{70.79$\pm$0.49}$^{\dagger}$ &
58.63$\pm$0.76$^{\dagger}$ & \textbf{78.62$\pm$0.55}$^{\dagger}$
\\

+ \uniform (Online) &
48.39$\pm$0.62 & \textbf{67.86$\pm$0.50}$^{\dagger}$ &
52.97$\pm$0.64$^{\dagger}$ & \textbf{70.63$\pm$0.50}$^{\dagger}$ &
\textbf{59.67$\pm$0.70}$^{\dagger}$ & \textbf{78.73$\pm$0.55}$^{\dagger}$
\\

\midrule

ProtoNet (cosine) &
\textbf{50.03$\pm$0.61} & 61.56$\pm$0.53 &
52.85$\pm$0.64 & 62.11$\pm$0.52 &
60.01$\pm$0.73 & 72.75$\pm$0.59
\\

+ \uniform (Offline) &
\textbf{50.07$\pm$0.59}$^{\dagger}$ & \textbf{66.33$\pm$0.52}$^{\dagger}$ &
\textbf{54.27$\pm$0.65}$^{\dagger}$ & \textbf{70.85$\pm$0.51}$^{\dagger}$ &
60.27$\pm$0.75$^{\dagger}$ & \textbf{78.36$\pm$0.54}$^{\dagger}$
\\

+ \uniform (Online) &
\textbf{50.06$\pm$0.61}$^{\dagger}$ & \textbf{65.99$\pm$0.52}$^{\dagger}$ &
\textbf{53.90$\pm$0.63}$^{\dagger}$ & 68.78$\pm$0.51$^{\dagger}$ &
\textbf{61.37$\pm$0.72}$^{\dagger}$ & 77.81$\pm$0.56$^{\dagger}$
\\

\midrule

MAML &
\textbf{46.88$\pm$0.60} & 55.16$\pm$0.55 &
49.92$\pm$0.65 & 63.93$\pm$0.59 &
55.37$\pm$0.74 & \textbf{72.93$\pm$0.60}
\\

+ \uniform (Offline) &
\textbf{46.67$\pm$0.63}$^{\dagger}$ & \textbf{62.09$\pm$0.55}$^{\dagger}$ &
\textbf{52.65$\pm$0.65}$^{\dagger}$ & \textbf{66.76$\pm$0.57}$^{\dagger}$ &
54.58$\pm$0.77 & 72.00$\pm$0.66
\\

+ \uniform (Online) &
\textbf{46.70$\pm$0.61}$^{\dagger}$ & \textbf{61.62$\pm$0.54}$^{\dagger}$ &
51.17$\pm$0.68$^{\dagger}$ & 65.63$\pm$0.57$^{\dagger}$ &
\textbf{57.15$\pm$0.74}$^{\dagger}$ & 71.67$\pm$0.67
\\

\midrule

ANIL &
\textbf{46.59$\pm$0.60} & \textbf{63.47$\pm$0.55} &
\textbf{49.65$\pm$0.65} & 59.51$\pm$0.56 &
\textbf{54.77$\pm$0.76} & 69.28$\pm$0.67
\\

+ \uniform (Offline) &
\textbf{46.93$\pm$0.62}$^{\dagger}$ & 62.75$\pm$0.60 &
\textbf{49.56$\pm$0.62}$^{\dagger}$ & \textbf{64.72$\pm$0.60}$^{\dagger}$ &
54.15$\pm$0.79$^{\dagger}$ & \textbf{70.44$\pm$0.69}$^{\dagger}$
\\

+ \uniform (Online) &
\textbf{46.82$\pm$0.63}$^{\dagger}$ & 62.63$\pm$0.59 &
\textbf{49.82$\pm$0.68}$^{\dagger}$ & \textbf{64.51$\pm$0.62}$^{\dagger}$ &
\textbf{55.18$\pm$0.74}$^{\dagger}$ & 69.55$\pm$0.71$^{\dagger}$
\\

\bottomrule
\end{tabular}
}
\vspace{-1em}
\end{table*}

\subsection{Online approximation of the proposal distribution}
\label{subsec:online}

Although the offline formulation is better suited for analysis experiments, it is expensive as it requires a pre-training phase for the proposal network and $2$ forward passes during episodic training (one for the episode loss, another for the proposal density).
In this subsection, we show that the online formulation faithfully approximates offline sampling and can retain most of the performance improvements from \uniform.
We take the same $24$ scenarios as in the previous subsection, and compare baseline sampling against offline and online \uniform.
\cref{tab:online_full} reports the full suite of results.

We observe that baseline is statistically competitive on $9/24$ scenarios; on the other hand, offline and online \uniform perform similarly on aggregate, as they are within the best results in $16/24$ and $15/24$ scenarios respectively.
Similar to its offline counterpart, online \uniform does better than or comparable to baseline sampling in $21$ out of $24$ scenarios.
On the $3/24$ scenarios where online \uniform underperforms compared to baseline, the average degradation is $-0.92\%$, and at most $-1.26\%$ (ignoring the standard deviations).
Conversely, in the remaining scenarios, it boosts accuracy by as much as $6.67\%$ and on average by $2.24\%$ (ignoring the standard deviations).
Therefore, using online \uniform, while computationally comparable to baseline sampling, results in a boost in few-shot performance; when it underperforms, it trails closely.
We also compute the mean accuracy difference between the offline and online formulation, which is $0.07\%\pm0.35$ accuracy points.
This confirms that both the offline and online methods produce quantitatively similar outcomes.

\begin{table*}[h]
\centering
\caption{
    \small
    \textbf{Few-shot accuracies for 5-way cross-domain few-shot episodes after training on \mimg.}
    The mean accuracy and the $95\%$ confidence interval are reported for evaluation done over $1$k test episodes.
    The first row in every scenario denotes baseline sampling.
    Best results for a fixed scenario are shown in bold.
}
\label{tab:cross_domain_slice}
\setlength{\tabcolsep}{3pt}
{
\small
\begin{tabular}{l cc cc}
\toprule

& \multicolumn{2}{c}{CUB-200} & \multicolumn{2}{c}{Describable Textures} \\
\cmidrule(lr){2-3} \cmidrule(lr){4-5}

& \multicolumn{2}{c}{\resnet} & \multicolumn{2}{c}{\cnn}   \\
\cmidrule(lr){2-3} \cmidrule(lr){4-5}

& 1-shot (\%) & 5-shot (\%) & 1-shot (\%) & 5-shot (\%) \\

\midrule

ProtoNet (cosine) &
38.67$\pm$0.60 & 49.75$\pm$0.57 & 32.09$\pm$0.45 & 38.44$\pm$0.41
\\

+ \uniform (Online) &
\textbf{40.55$\pm$0.60} & \textbf{56.30$\pm$0.55} & \textbf{33.63$\pm$0.47} & \textbf{43.28$\pm$0.44}
\\

\midrule

MAML &
35.80$\pm$0.56 & 45.16$\pm$0.62 & 29.47$\pm$0.46 & 37.85$\pm$0.47
\\

+ \uniform (Online) &
\textbf{37.18$\pm$0.55} & \textbf{46.58$\pm$0.58} & \textbf{31.84$\pm$0.49} & \textbf{40.81$\pm$0.44}
\\

\bottomrule
\end{tabular}
}
\vspace{-1em}
\end{table*}

\subsection{Better sampling improves cross-domain transfer}
\label{subsec:cross_domain}

To further validate the role of episode sampling as a way to improve generalization, we evaluate the models trained in the previous subsection on episodes from completely different domains.
Specifically, we use the models trained on \mimg with baseline and online \uniform sampling to evaluate on the test episodes of CUB-200~\citep{welinder2010caltech}, Describable Textures~\citep{cimpoi2014describing}, FGVC Aircrafts~\citep{maji2013fine}, and VGG Flowers~\citep{nilsback2006visual}, following the splits of~\citet{Triantafillou2020Meta-Dataset:}.
\cref{tab:cross_domain_slice} displays results for ProtoNet (cosine) and MAML on CUB-200 and Describable Textures, with the complete set of experiments available in~\cref{app:cross_domain}.
Out of the 64 total cross-domain scenarios, online \uniform does statistically better in $49/64$ scenarios, comparable in $12/64$ scenarios and worse in only $3/64$ scenarios.
These results further go to show that sampling matters in episodic training.

\begin{table*}[h]
\centering
\caption{
    \small
    \textbf{Few-shot accuracies on benchmark datasets for 5-way few-shot episodes using FEAT.}
    The mean accuracy and the $95\%$ confidence interval are reported for evaluation done over $10$k test episodes with a \resnet.
    The first row in every scenario denotes baseline sampling.
    Best results for a fixed scenario are shown in bold.
    \uniform (Online) improves FEAT's accuracy in $3/4$ scenarios, demonstrating that sampling matters even for state-of-the-art few-shot methods.
}
\label{tab:sota}
\setlength{\tabcolsep}{3pt}
{
\small
\begin{tabular}{l cc cc}
\toprule

& \multicolumn{2}{c}{\mimg} & \multicolumn{2}{c}{\timg} \\
\cmidrule(lr){2-3} \cmidrule(lr){4-5}

& 1-shot (\%) & 5-shot (\%) & 1-shot (\%) & 5-shot (\%) \\

\midrule

FEAT
& 66.02$\pm$0.20 & 81.17$\pm$0.14 &
\textbf{70.50$\pm$0.23} & 84.26$\pm$0.16
\\

+ \uniform (Online)
& \textbf{66.27$\pm$0.20} & \textbf{81.54$\pm$0.14}
& \textbf{70.61$\pm$0.23} & \textbf{84.42$\pm$0.16}
\\

\bottomrule
\end{tabular}
}
\vspace{-1.5em}
\end{table*}

\subsection{Better sampling improves few-shot classification}

The results in the previous subsections suggest that online \uniform yields a \emph{simple and universally applicable} method to improve episode sampling.
To validate that state-of-the-art methods can also benefit from better sampling, we take the recently proposed FEAT~\citep{ye2020fewshot} algorithm and augment it with our IS-based implementation of online \uniform.
Concretely, we use their open-source implementation\footnote{Available at: \url{https://github.com/Sha-Lab/FEAT}} to train both baseline and online \uniform sampling.
We use the prescribed hyper-parameters without any modifications.
Results for \resnet on \mimg and \timg are reported in~\cref{tab:sota}, where online \uniform outperforms baseline sampling on $3/4$ scenarios and is matched on the remaining one.
Thus, better episodic sampling can improve few-shot classification even for the very best methods.

\section{Conclusion}
\label{sec:conclusion}

This manuscript presents a careful study of sampling in the context of few-shot learning, with an eye on episodes and their difficulty.
Following an empirical study of episode difficulty, we propose an importance sampling-based method to compare different episode sampling schemes.
Our experiments suggest that sampling uniformly over episode difficulty performs best across datasets, training algorithms, network architectures and few-shot protocols.
Avenues for future work include devising better sampling strategies, analysis beyond few-shot classification (\emph{e.g.}, regression, reinforcement learning), and a theoretical grounding explaining our observations.

\clearpage
\pagebreak

{
\small
\bibliographystyle{abbrvnat}
\bibliography{bibliography}
}

\clearpage
\pagebreak

\appendix

\section{Experimental setup}
\label{app:setup}

\subsection{Datasets}

We use two standardized few-shot image classification datasets.

\emph{\mimg}:
This dataset~\citep{vinyals2016matching} is a subset of ImageNet~\citep{deng2009imagenet} and consists of $64$ classes for training, $16$ for validation, and $20$ for testing.
There are $600$ images per class, with images of size $84 \times 84$.
Multiple versions of this dataset exist in the literature; we use the version by~\citet{ravi2016optimization}.

\emph{\timg}:
A larger subset of ImageNet, \timg~\citep{ren2018meta} consists of $608$ classes split into $351$, $97$, and $160$ for training, validation, and testing, respectively.
Each class has about $1,300$ images of size $84 \times 84$.
This dataset ensures that the train, validation, and test classes do not have any semantic overlap and is proposed as a harder few-shot learning benchmark.

We also use the test splits of the following four datasets, as defined by~\citet{Triantafillou2020Meta-Dataset:}.

\emph{CUB-200}:
CUB-200 was collected by~\citet{welinder2010caltech} and contains $6,033$ bird images classified into $200$ bird species.
The original version of the dataset contains $43$ images that are also present in ImageNet.
We remove these duplicates to avoid overestimating the transfer capability during evaluation.
The test split contains $30$ classes.

\emph{Describable Textures}:
Proposed by~\citet{cimpoi2014describing}, the task of this dataset is to classify images into $47$ texture classes.
Each of the $5,640$ images ($120$ samples per class) contains at least $90\%$ of the class' texture, with sizes between $300 \times 300$ and $640 \times 640$ pixels.
The train split has $33$ classes, while validation and test splits both consist of $7$ classes.

\emph{VGG Flowers}:
Originally introduced by~\citet{nilsback2006visual}, VGG Flowers consists of $102$ flower categories with each category containing between $40$ and $258$ images.
While we use~\citet{Triantafillou2020Meta-Dataset:}'s train ($71$ classes), validation ($15$ classes), and test ($16$ classes) splits, our models operate on the raw images, not the cropped versions.

\emph{FGVC Aircrafts}:
\citet{maji2013fine} introduced this dataset containing $10,200$ images of aircraft partitioned into $102$ classes, each with $100$ samples.
The test split contains $15$ classes.
As for VGG Flowers, we do not crop those images using bounding box information, thus increasing the classification difficulty.

\subsection{Network architectures}

We train two of the most popular network architectures in few-shot learning literature.

\emph{\cnn}:
This architecture~\citep{vinyals2016matching} consists of $4$ convolutional layers with $64$ channels per layer.

\emph{\resnet}:
From the family of deep residual networks~\citep{he2016deep}, this architecture has $4$ blocks, each block constituting $3$ convolutional layers with $64 \times 2^{l - 1}$ channels per layer in the $l$'th block.
Two versions of this network architecture exist in the literature; we use the one by~\citet{oreshkin2018tadam}.
The other version by~\citet{lee2019meta} is $1.25 \times$ wider and has more parameters.

Both architectures use batch normalization~\citep{ioffe2015batch} after every convolutional layer with ReLU as the non-linearity.
We do not use dropout~\citep{srivastava2014dropout} or any of its variants, like~\citet{ghiasi2018dropblock}.
For MAML and ANIL, a fully-connected layer is appended at the top of the networks.

\subsection{Training algorithms}

For the metric-based family, we use ProtoNet with Euclidean~\citep{snell2017prototypical} and scaled negative cosine similarity measures~\citep{gidaris2018dynamic}.
Based on the implementation of~\citet{gidaris2018dynamic}, we add a learnable parameter that scales the cosine similarity.
Additionally, we use MAML~\citep{finn2017model} and ANIL~\citep{raghu2019rapid} as representative gradient-based algorithms.
We use the open-source library \texttt{lear2learn}~\citep{Arnold2020-ss}\footnote{Available at: \url{https://github.com/learnables/learn2learn}} to implement these algorithms.

\subsection{Sampling methods}

We compare four sampling methods -- \easy, \hard, \curriculum, and \uniform.
In each case, we mimic the target distribution using importance sampling (\cref{subsec:target_distribution}).

We also have baseline sampling in our comparisons.
This involves episodic training without the use of any weighting techniques, hence sampling episodes from the distribution $q(\tau)$ without making any changes to it (\cref{subsec:episodic_sampling}).
This is the default sampling strategy for few-shot episodic training.

\subsection{Hyper-parameters}

We tune hyper-parameters for each algorithm and dataset to work well across different few-shot settings and network architectures.
Additionally, we keep the hyper-parameters the same across all different sampling methods for a fair comparison.

All models are trained using ADAM~\citep{kingma2015adam} with a learning rate of $10^{-3}$ on a single NVIDIA Tesla V100 GPU.
MAML and ANIL use an adaptation learning rate of $0.01$ and $0.1$ respectively, with $5$ adaptation steps taken in both cases.
All models are trained for a total of $20$k iterations, with a mini-batch of size $16$ and $32$ for \mimg and \timg respectively.
After every $1$k iterations, we evaluate on $1$k validation episodes.
The model with the best validation performance is finally evaluated on $1$k test episodes.

\section{Episode difficulty is approximately normally distributed}

\begin{figure*}[t]
    \begin{center}
        \includegraphics[width=1.\linewidth]{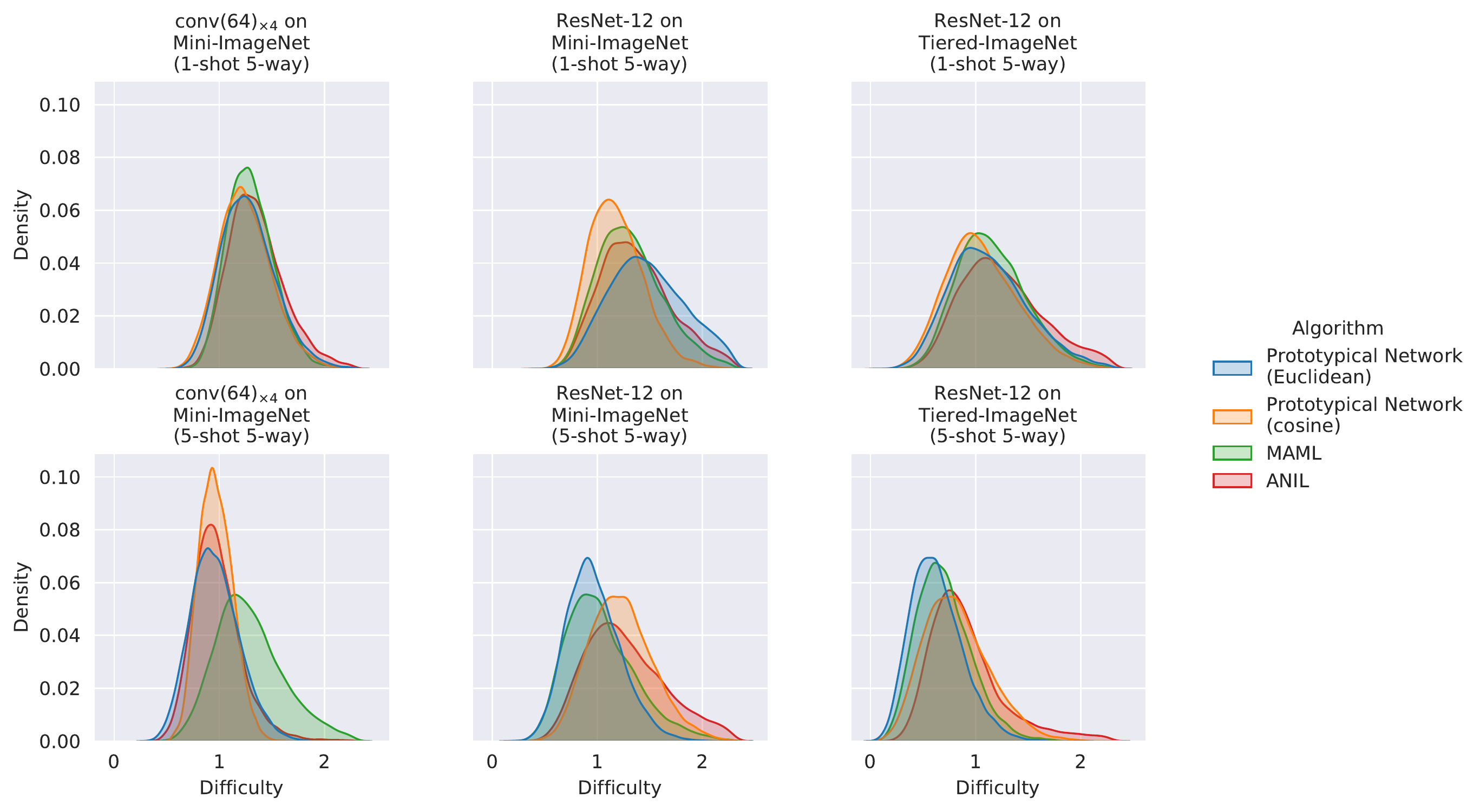}
    \end{center}
    \caption{
        \small
        \textbf{Episode difficulty is approximately normally distributed - density plots.}
        Density plots of the episode difficulty computed by \cnn's on \mimg (left), \resnet's on \mimg (center) and \resnet's on \timg (right), trained using ProtoNets (Euclidean and cosine), MAML and ANIL (depicted in the legend).
        The values are computed over $10$k test episodes.
        The top row is for \osfw episodes and the bottom row is for \fsfw episodes.
        All the plots follow a bell curve, with the density peak in the middle, which quickly drops-off on either side of the peak.
    }
    \label{fig:hardness_normal_full}
\end{figure*}

\begin{figure*}[t]
    \begin{center}
        \includegraphics[width=1.\linewidth]{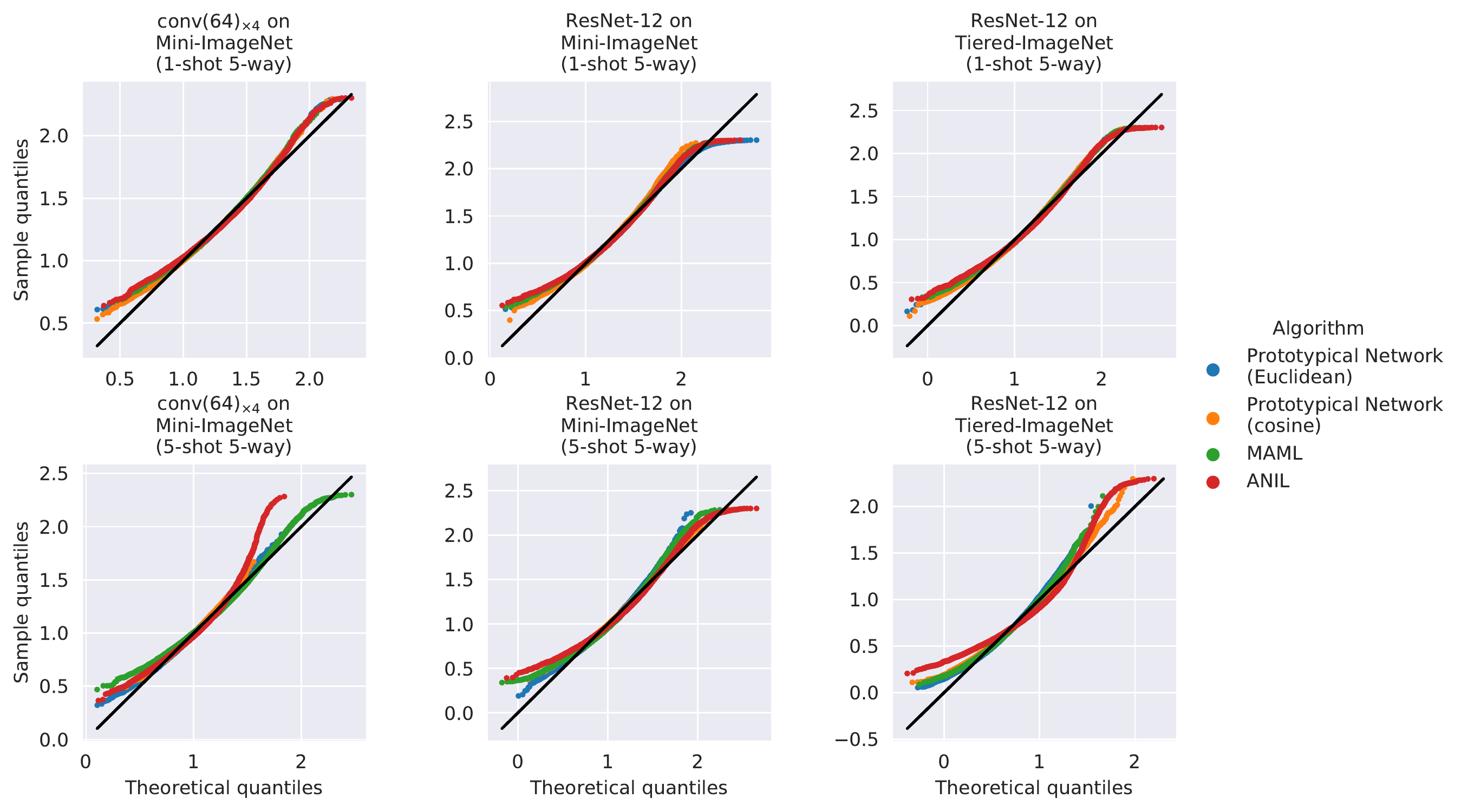}
    \end{center}
    \caption{
        \small
        \textbf{Episode difficulty is approximately normally distributed - Q-Q plots.}
        Q-Q plots of the episode difficulty computed by \cnn's on \mimg (left), \resnet's on \mimg (center) and \resnet's on \timg (right), trained using ProtoNets (Euclidean and cosine), MAML and ANIL (depicted in the legend).
        The values are computed over $10$k test episodes and are plotted against normal distributions with the same mean and standard deviation as the episode difficulties.
        The top row is for \osfw episodes and the bottom row is for \fsfw episodes.
        We also include the identity line in each plot (in black).
        The closer the curve is to the identity line, the closer the distribution is to a normal.
    }
    \label{fig:hardness_normal_qq_full}
\end{figure*}

Sampling episodes from $q(\tau)$ (\cref{subsec:episodic_sampling}) induces a distribution over their difficulty $\Omega_{l_\theta}$.
Our proposed method estimates this as a normal distribution (\cref{subsec:proposal_distribution}), and here we justify why.

We train \cnn's on \mimg and \resnet's on both \mimg and \timg using baseline sampling.
This is done using all four learning algorithms -- ProtoNet (Euclidean and cosine), MAML and ANIL -- for \osfw and \fsfw classification.
We compute the episode difficulty over $10$k test episodes, sampled using the episode distribution $q(\tau)$.

\cref{fig:hardness_normal_full} illustrates the density plots of the computed difficulties.
We observe that the episode difficulties follow a bell curve in each case, which is naturally modeled with a normal distribution.
\cref{fig:hardness_normal_qq_full} includes Q-Q plots for the same, plotted against normal distributions with the same mean and standard deviation as the corresponding episode difficulties.
These plots are typically used to assess normality -- the closer the curve is to the identity line, the closer the distribution is to a normal, which is observed here.

\begin{table*}[ht]
    \centering
    \caption{
        \small
        \textbf{Episode difficulty is approximately normally distributed - Shapiro-Wilk normality tests.}
        We compute the episode difficulty for different datasets, algorithms and network architectures, for both the \osfw and \fsfw settings.
        This is done for $10$k test episodes each.
        In each case, we subsample $50$ values $100$ times and run the Shapiro-Wilk test on these subsets (with $\alpha = 0.05$).
        The rejection rates of the null hypothesis are averaged over everything but the axes mentioned in the left column and are mentioned in the column on the right.
        The average rejection rate does not exceed $20\%$.
    }
    \label{tab:shapiro_wilk}
    \setlength{\tabcolsep}{3pt}
    {
    \small
    \begin{tabular}{llc}
    \toprule
    & & Rejection rate (\%) \\
    \midrule
    \multirow{2}{*}{Dataset} & \mimg & 14.25 \\
    & \timg & 17.38 \\
    \midrule
    \multirow{2}{*}{Shots} & 1-shot & 14.17 \\
    & 5-shot & 16.42 \\
    \midrule
    \multirow{4}{*}{Algorithm} & ProtoNet (Euclidean) & 19.67 \\
    & ProtoNet (cosine) & 09.17 \\
    & MAML & 13.33 \\
    & ANIL & 19.00 \\ 
    \midrule
    \multirow{2}{*}{Network Architecture} & \cnn & 09.63 \\
    & \resnet & 18.13 \\
    \bottomrule
    \end{tabular}
    }
\end{table*}

We additionally run the Shapiro-Wilk test for normality~\citep{shapiro1965analysis} on the computed episode difficulties, which tests for the null hypothesis that the data is drawn from a normal distribution.
The p-value for this test is sensitive to the sample size -- for large sample sizes, trivial departures from the normal distribution can be detected, making the p-values unreliable.
Instead, we subsample $50$ values $100$ times and run the test on these subsets (with $\alpha = 0.05$).
\cref{tab:shapiro_wilk} summarizes the rejection rates of the null hypothesis averaged over datasets, shots, algorithms and network architectures.
Regardless of which axis the rejection rate is averaged over, it does not exceed $20\%$.
These results suggest that our assumption of estimating the induced distribution over the episode difficulty as a normal distribution is plausible.

\section{Episode difficulty is independent from modeling choices}

This section provides the full version of the figures from~\cref{subsubsec:episode_hardness_model}.

\begin{figure*}[ht]
    \begin{center}
        \includegraphics[width=1.\linewidth]{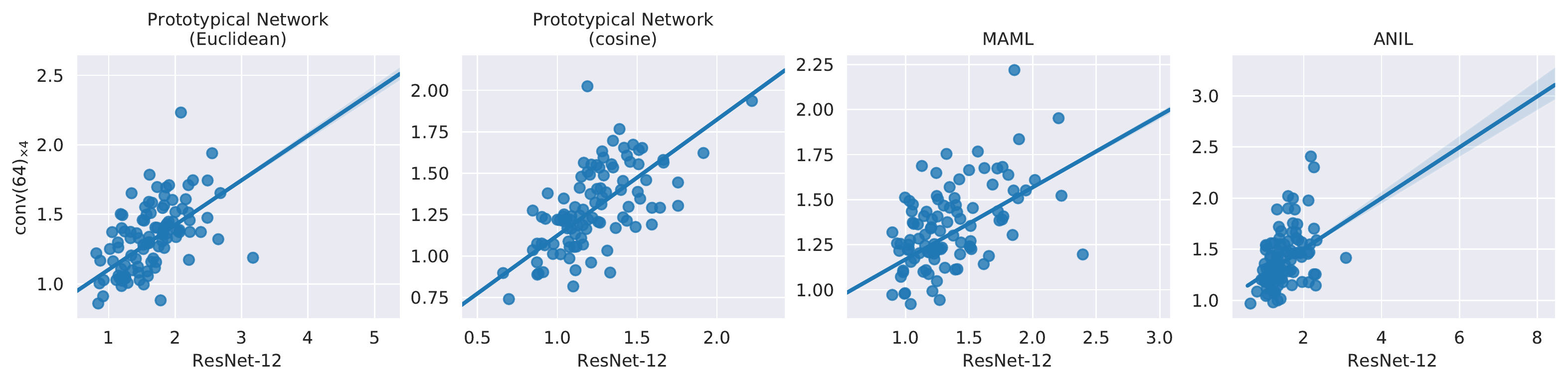}
    \end{center}
    \caption{
        \small
        \textbf{Episode difficulty transfers across network architectures.}
        Scatter plots (with regression line and $95\%$ confidence interval) of episode difficulties computed on $10$k \osfw \mimg episodes by \cnn's and \resnet's trained using different training algorithms.
        Similar to~\cref{fig:hardness_model}, we observe that an episode that is hard for one architecture is very likely to be hard for another, for all training algorithms.
    }
    \label{fig:hardness_model_full}
\end{figure*}

\cref{fig:hardness_model_full} reports correlation plots for the difficulty of episodes when measured with two different architectures.
We use \cnn and \resnet's trained on \mimg (\osfw) with all training algorithms to compute episode difficulties for $10$k test episodes.
We then compute the Spearman rank-order correlation coefficients across the two architectures, for a given algorithm.
The correlation coefficients are $0.57$ for ProtoNet (Euclidean), $0.72$ for ProtoNet (cosine), $0.58$ for MAML, and $0.49$ for ANIL.
As mentioned in~\cref{subsubsec:episode_hardness_model}, this positive correlation suggests that episode difficulty is transferred across network architectures with high probability.

\begin{figure*}[ht]
    \begin{center}
        \includegraphics[width=1.\linewidth]{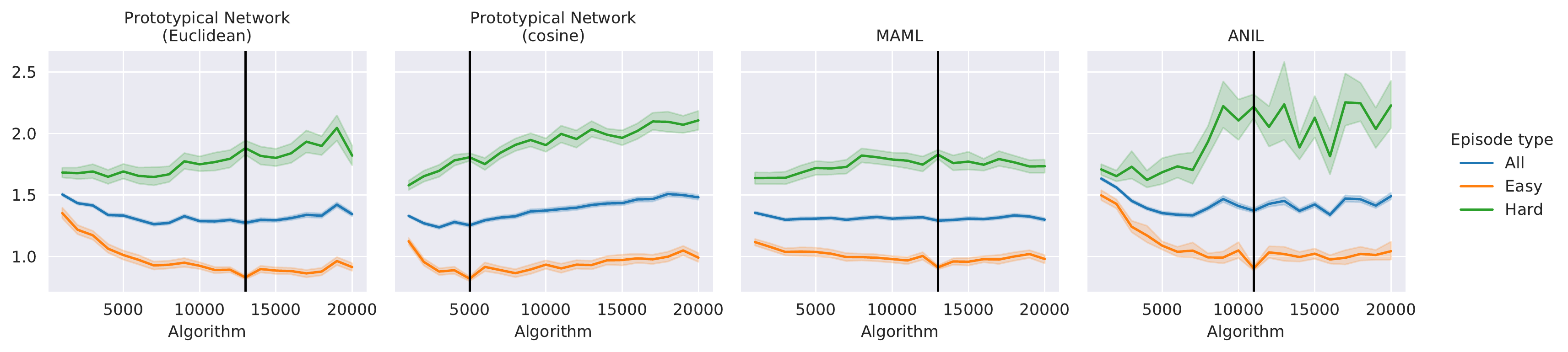}
    \end{center}
    \caption{
        \small
        \textbf{Episode difficulty transfers across model parameters during training.}
        Similar to~\cref{fig:hardness_time}, out of $1$k test episodes, we select the $50$ easiest and $50$ hardest episodes and track the difficulties of them all throughout training.
        The average difficulty of the episodes decreases over time, until convergence (vertical line), after which the model overfits.
        Additionally, easier episodes remain easy while harder episodes remain hard, indicating that episode difficulty transfers from one set of parameters to the next.
        Note that since we validate the models every $1$k iterations, these plots are not continuous and do not contain the values for the first $1$k training iterations (during which the episode difficulty drops quickly).
    }
    \label{fig:hardness_time_full}
\end{figure*}

\cref{fig:hardness_time_full} tracks the difficulty of easy and hard episodes over training iterations, for all four training algorithms.
Out of $1$k test episodes, we select the $50$ \emph{easiest} and $50$ \emph{hardest} episodes, \ie, episodes with the lowest and highest difficulties respectively.
We measure difficulty on these episodes every $1$k training iterations, and observe that the difficulty lines for easy and hard episodes never cross -- easy episodes remain easy and hard episodes remain hard.
This observation suggests that episode difficulty transfers across different model parameters during training, justifying our online estimation of difficulty parameters $\mu$ and $\sigma^2$ (\cref{subsec:proposal_distribution}).

\section{Comparing episode sampling methods}
\label{app:offline}

\begin{table*}[t]
\centering
\caption{
    \small
    \textbf{Few-shot accuracies on benchmark datasets for 5-way few-shot episodes in the offline setting.}
    The mean accuracy and the $95\%$ confidence interval are reported for evaluation done over $1$k test episodes.
    The first row in every scenario denotes baseline sampling.
    Best results for a fixed scenario are shown in bold.
    Results where a sampling technique is better than or comparable to baseline sampling are denoted by $\dagger$.
    Overall, \uniform is among the best sampling methods in $19/24$ scenarios.
}
\label{tab:offline_full}
\setlength{\tabcolsep}{2pt}
{
\small
\begin{tabular}{l cc cc cc}
\toprule

& \multicolumn{4}{c}{\mimg} & \multicolumn{2}{c}{\timg} \\
\cmidrule(lr){2-5} \cmidrule(lr){6-7}

& \multicolumn{2}{c}{\cnn}  & \multicolumn{2}{c}{\resnet} & \multicolumn{2}{c}{\resnet} \\
\cmidrule(lr){2-3} \cmidrule(lr){4-5} \cmidrule(lr){6-7}

& 1-shot (\%) & 5-shot (\%) & 1-shot (\%) & 5-shot (\%) & 1-shot (\%) & 5-shot (\%) \\

\midrule

ProtoNet (Euclidean) & \textbf{49.06$\pm$0.60} & 65.28$\pm$0.52 & 49.67$\pm$0.64 & 67.45$\pm$0.51 & \textbf{59.10$\pm$0.73} & 76.95$\pm$0.56 \\

+ \easy & \textbf{48.83$\pm$0.61}$^{\dagger}$ & 65.92$\pm$0.55$^{\dagger}$ & 51.08$\pm$0.63$^{\dagger}$ & 67.30$\pm$0.52$^{\dagger}$ & 57.68$\pm$0.75 & \textbf{78.10$\pm$0.53}$^{\dagger}$ \\

+ \hard & 45.69$\pm$0.61 & \textbf{66.47$\pm$0.52}$^{\dagger}$ & 52.50$\pm$0.62$^{\dagger}$ & \textbf{71.03$\pm$0.51}$^{\dagger}$ & 54.85$\pm$0.71 & 76.15$\pm$0.56 \\

+ \curriculum & 48.23$\pm$0.63 & 65.77$\pm$0.51$^{\dagger}$ & 50.00$\pm$0.61$^{\dagger}$ & 70.49$\pm$0.51$^{\dagger}$ & \textbf{59.15$\pm$0.76}$^{\dagger}$ & \textbf{78.25$\pm$0.53}$^{\dagger}$ \\

+ \uniform & 48.19$\pm$0.62 & \textbf{66.73$\pm$0.52}$^{\dagger}$ & \textbf{53.94$\pm$0.63}$^{\dagger}$ & \textbf{70.79$\pm$0.49}$^{\dagger}$ & \textbf{58.63$\pm$0.76}$^{\dagger}$ & \textbf{78.62$\pm$0.55}$^{\dagger}$ \\

\midrule

ProtoNet (cosine) & \textbf{50.03$\pm$0.61} & 61.56$\pm$0.53 & 52.85$\pm$0.64 & 62.11$\pm$0.52 & \textbf{60.01$\pm$0.73} & 72.75$\pm$0.59 \\

+ \easy & \textbf{49.60$\pm$0.61}$^{\dagger}$ & 65.17$\pm$0.53$^{\dagger}$ & 53.35$\pm$0.63$^{\dagger}$ & 63.55$\pm$0.53$^{\dagger}$ & \textbf{60.03$\pm$0.75}$^{\dagger}$ & 74.65$\pm$0.57$^{\dagger}$ \\

+ \hard & 49.01$\pm$0.60 & \textbf{66.45$\pm$0.50}$^{\dagger}$ & 52.65$\pm$0.63$^{\dagger}$ & 70.15$\pm$0.51$^{\dagger}$ & 55.44$\pm$0.72 & 75.97$\pm$0.55$^{\dagger}$ \\

+ \curriculum & 49.38$\pm$0.61 & 64.12$\pm$0.53$^{\dagger}$ & 53.21$\pm$0.65$^{\dagger}$ & 65.89$\pm$0.52$^{\dagger}$ & \textbf{60.37$\pm$0.76}$^{\dagger}$ & 75.32$\pm$0.58$^{\dagger}$ \\

+ \uniform & \textbf{50.07$\pm$0.59}$^{\dagger}$ & \textbf{66.33$\pm$0.52}$^{\dagger}$ & \textbf{54.27$\pm$0.65}$^{\dagger}$ & \textbf{70.85$\pm$0.51}$^{\dagger}$ & \textbf{60.27$\pm$0.75}$^{\dagger}$ & \textbf{78.36$\pm$0.54}$^{\dagger}$ \\

\midrule

MAML & \textbf{46.88$\pm$0.60} & 55.16$\pm$0.55 & 49.92$\pm$0.65 & 63.93$\pm$0.59 & \textbf{55.37$\pm$0.74} & \textbf{72.93$\pm$0.60} \\

+ \easy & 44.52$\pm$0.60 & 57.36$\pm$0.59$^{\dagger}$ & 51.62$\pm$0.67$^{\dagger}$ & 64.33$\pm$0.61$^{\dagger}$ & 53.39$\pm$0.79 & 69.81$\pm$0.68 \\

+ \hard & 42.93$\pm$0.61 & 60.42$\pm$0.55$^{\dagger}$ & 49.57$\pm$0.69$^{\dagger}$ & \textbf{66.93$\pm$0.55}$^{\dagger}$ & 50.48$\pm$0.73 & 71.20$\pm$0.63 \\

+ \curriculum & 45.42$\pm$0.60 & \textbf{61.61$\pm$0.55}$^{\dagger}$ & \textbf{52.21$\pm$0.67}$^{\dagger}$ & 66.25$\pm$0.60$^{\dagger}$ & 54.13$\pm$0.77 & 71.47$\pm$0.63 \\

+ \uniform & \textbf{46.67$\pm$0.63}$^{\dagger}$ & \textbf{62.09$\pm$0.55}$^{\dagger}$ & \textbf{52.65$\pm$0.65}$^{\dagger}$ & \textbf{66.76$\pm$0.57}$^{\dagger}$ & 54.58$\pm$0.77 & 72.00$\pm$0.66 \\

\midrule

ANIL & \textbf{46.59$\pm$0.60} & \textbf{63.47$\pm$0.55} & \textbf{49.65$\pm$0.65} & 59.51$\pm$0.56 & 54.77$\pm$0.76 & 69.28$\pm$0.67 \\

+ \easy & 44.83$\pm$0.63 & 62.23$\pm$0.56 & 49.40$\pm$0.64$^{\dagger}$ & 56.73$\pm$0.60 & 54.50$\pm$0.80$^{\dagger}$ & 65.45$\pm$0.66 \\

+ \hard & 43.30$\pm$0.58 & 59.87$\pm$0.55 & 47.91$\pm$0.62 & 62.05$\pm$0.59$^{\dagger}$ & 50.22$\pm$0.71 & 62.06$\pm$0.65 \\

+ \curriculum & 45.69$\pm$0.60 & \textbf{63.00$\pm$0.54}$^{\dagger}$ & \textbf{50.22$\pm$0.66}$^{\dagger}$ & 61.76$\pm$0.57$^{\dagger}$ & \textbf{55.59$\pm$0.78}$^{\dagger}$ & \textbf{69.83$\pm$0.73}$^{\dagger}$ \\

+ \uniform & \textbf{46.93$\pm$0.62}$^{\dagger}$ & 62.75$\pm$0.60 & \textbf{49.56$\pm$0.62}$^{\dagger}$ & \textbf{64.72$\pm$0.60}$^{\dagger}$ & 54.15$\pm$0.79$^{\dagger}$ & \textbf{70.44$\pm$0.69}$^{\dagger}$ \\

\bottomrule
\end{tabular}
}
\end{table*}

In addition to the discussion in~\cref{subsec:offline}, this section presents the full suite of results for the comparison of different episode sampling methods.
We compute results over $2$ datasets, $2$ network architectures, $4$ algorithms and $2$ few-shot protocols, resulting in $24$ total scenarios.
\cref{tab:offline_full} contains all performance numbers.
As mentioned in the main text, \uniform is among the better sampling schemes in $19/24$ scenarios, followed by baseline sampling which is competitive in $10/24$ scenarios.
Importantly, when \uniform underperforms it is a close second.

\section{Difference in effectiveness in the 1- and 5-shot settings}

The $1$-shot setting is inherently noisier than $5$-shot.
Support samples are randomly drawn from the class-populations, which are then used to construct the few-shot classifier.
Sampling only $1$ support per-class is more susceptible to outliers in the query set than sampling $5$ (the higher the support-shot, the better the estimate of the class-population).
This noise propagates to the loss (in the case of baseline sampling) as well as the weighted loss (in the case of \uniform sampling).
Hence, larger noise degrades the approximation to a uniform distribution over episode difficulty and ultimately results in \uniform not getting as much gain in the $1$-shot setting.

We empirically confirm this hypothesis.
We use the same $24$ scenarios as the ones in~\cref{subsec:offline,subsec:online} and compare the training procedures of \uniform under $1$- vs. $5$-shot settings.
Using~\cref{eq:mc-ess}, we compute the per-episode weighted loss during the training process, followed by the per-mini-batch standard deviation.
The average deviation is higher under the $1$-shot than the $5$-shot setting in all scenarios (for both offline and online settings).
Additionally, the average deviation is $\approx 1.9$ times larger under the $1$-shot setting.
These experiments confirm the above hypothesis and help explain why \uniform (online) outperforms the baseline in (only) $5/12$ $1$-shot scenarios, is comparable in $6/12$, and underperforms in $1/12$.

\begin{table*}[h]
\centering
\caption{
    \small
    \textbf{Few-shot accuracies on benchmark datasets after training on \mimg for 5-way few-shot episodes in the offline and online settings.}
    The mean accuracy and the $95\%$ confidence interval are reported for evaluation done over $1,000$ test episodes.
    Best results for a fixed scenario are shown in bold.
    The first row in every scenario denotes baseline sampling.
    Compared to baseline sampling, online \uniform does statistically better in $49/64$ scenarios, comparable in $12/64$ scenarios and worse in only $3/64$ scenarios.
}
\label{tab:cross_domain}
\setlength{\tabcolsep}{3pt}
{
\small
\begin{tabular}{l cc cc}
\toprule

& \multicolumn{2}{c}{\cnn}  & \multicolumn{2}{c}{\resnet} \\
\cmidrule(lr){2-3} \cmidrule(lr){4-5}

& 1-shot (\%) & 5-shot (\%) & 1-shot (\%) & 5-shot (\%) \\

\midrule

& \multicolumn{4}{c}{CUB-200} \\
\cmidrule(lr){2-5}

ProtoNet (Euclidean) &
\textbf{37.24$\pm$0.53} & 52.07$\pm$0.53 & 36.53$\pm$0.54 & 51.49$\pm$0.56
\\

+ \uniform (Online) &
\textbf{37.08$\pm$0.53} & \textbf{53.32$\pm$0.53} & \textbf{39.48$\pm$0.56} & \textbf{56.57$\pm$0.55}
\\

\midrule

ProtoNet (cosine) &
37.49$\pm$0.54 & 49.31$\pm$0.53 & 38.67$\pm$0.60 & 49.75$\pm$0.57
\\

+ \uniform (Online) &
\textbf{41.56$\pm$0.58} & \textbf{54.17$\pm$0.53} & \textbf{40.55$\pm$0.60} & \textbf{56.30$\pm$0.55}
\\

\midrule

MAML &
34.52$\pm$0.53 & \textbf{47.11$\pm$0.60} & 35.80$\pm$0.56 & 45.16$\pm$0.62
\\

+ \uniform (Online) &
\textbf{35.84$\pm$0.54} & \textbf{46.67$\pm$0.55} & \textbf{37.18$\pm$0.55} & \textbf{46.58$\pm$0.58}
\\

\midrule

ANIL &
35.40$\pm$0.54 & 38.20$\pm$0.56 & 33.20$\pm$0.54 & 39.26$\pm$0.58
\\

+ \uniform (Online) &
\textbf{36.89$\pm$0.55} & \textbf{42.83$\pm$0.58} & \textbf{34.47$\pm$0.56} & \textbf{42.08$\pm$0.58}
\\

\midrule

& \multicolumn{4}{c}{Describable Textures} \\
\cmidrule(lr){2-5}

ProtoNet (Euclidean) &
32.05$\pm$0.45 & \textbf{45.03$\pm$0.44} & 31.87$\pm$0.45 & 44.10$\pm$0.43
\\

+ \uniform (Online) &
\textbf{32.69$\pm$0.49} & \textbf{45.23$\pm$0.43} & \textbf{33.55$\pm$0.46} & \textbf{47.37$\pm$0.43}
\\

\midrule

ProtoNet (cosine) &
32.09$\pm$0.45 & 38.44$\pm$0.41 & 31.48$\pm$0.45 & 39.46$\pm$0.41
\\

+ \uniform (Online) &
\textbf{33.63$\pm$0.47} & \textbf{43.28$\pm$0.44} & \textbf{32.69$\pm$0.48} & \textbf{45.56$\pm$0.42}
\\

\midrule

MAML &
29.47$\pm$0.46 & 37.85$\pm$0.47 & \textbf{32.19$\pm$0.48} & 41.14$\pm$0.46
\\

+ \uniform (Online) &
\textbf{31.84$\pm$0.49} & \textbf{40.81$\pm$0.44} & 31.65$\pm$0.46 & \textbf{43.21$\pm$0.44}
\\

\midrule

ANIL &
29.86$\pm$0.46 & 40.69$\pm$0.46 & 28.85$\pm$0.41 & 37.04$\pm$0.44
\\

+ \uniform (Online) &
\textbf{31.29$\pm$0.48} & \textbf{41.42$\pm$0.45} & \textbf{31.38$\pm$0.47} & \textbf{39.03$\pm$0.47}
\\

\midrule

& \multicolumn{4}{c}{FGVC-Aircraft} \\
\cmidrule(lr){2-5}

ProtoNet (Euclidean) &
\textbf{26.03$\pm$0.37} & 39.41$\pm$0.48 & 25.98$\pm$0.39 & 36.76$\pm$0.45
\\

+ \uniform (Online) &
\textbf{26.18$\pm$0.38} & \textbf{40.23$\pm$0.46} & \textbf{27.43$\pm$0.42} & \textbf{38.49$\pm$0.46}
\\

\midrule

ProtoNet (cosine) &
\textbf{27.11$\pm$0.39} & 32.14$\pm$0.38 & 25.23$\pm$0.39 & 32.07$\pm$0.41
\\

+ \uniform (Online) &
\textbf{27.15$\pm$0.38} & \textbf{37.78$\pm$0.45} & \textbf{26.89$\pm$0.39} & \textbf{37.42$\pm$0.44}
\\

\midrule

MAML &
\textbf{26.78$\pm$0.38} & \textbf{34.21$\pm$0.41} & 25.50$\pm$0.39 & 29.38$\pm$0.40
\\

+ \uniform (Online) &
\textbf{26.62$\pm$0.39} & \textbf{34.41$\pm$0.44} & \textbf{26.22$\pm$0.39} & \textbf{30.21$\pm$0.43}
\\

\midrule

ANIL &
\textbf{25.67$\pm$0.37} & 27.17$\pm$0.36 & 23.27$\pm$0.31 & 24.52$\pm$0.29
\\

+ \uniform (Online) &
\textbf{25.60$\pm$0.37} & \textbf{27.92$\pm$0.39} & \textbf{23.78$\pm$0.34} & \textbf{28.70$\pm$0.39}
\\

\midrule

& \multicolumn{4}{c}{VGG Flowers} \\
\cmidrule(lr){2-5}

ProtoNet (Euclidean) &
53.50$\pm$0.63 & 70.96$\pm$0.51 & \textbf{57.74$\pm$0.68} & 74.87$\pm$0.49
    \\

+ \uniform (Online) &
\textbf{54.72$\pm$0.65} & \textbf{73.59$\pm$0.49} & 55.94$\pm$0.67 & \textbf{76.62$\pm$0.50}
\\

\midrule

ProtoNet (cosine) &
52.94$\pm$0.62 & 66.04$\pm$0.53 & 52.98$\pm$0.65 & 66.79$\pm$0.51
\\

+ \uniform (Online) &
\textbf{54.23$\pm$0.63} & \textbf{71.93$\pm$0.48} & \textbf{57.06$\pm$0.65} & \textbf{67.31$\pm$0.48}
\\

\midrule

MAML &
\textbf{49.70$\pm$0.60} & \textbf{63.69$\pm$0.54} & \textbf{50.13$\pm$0.64} & 61.41$\pm$0.63
\\

+ \uniform (Online) &
\textbf{49.72$\pm$0.60} & \textbf{63.52$\pm$0.54} & \textbf{49.53$\pm$0.65} & \textbf{63.99$\pm$0.58}
\\

\midrule

ANIL &
\textbf{47.03$\pm$0.65} & 46.40$\pm$0.66 & \textbf{42.05$\pm$0.67} & 40.01$\pm$0.65
\\

+ \uniform (Online) &
\textbf{47.48$\pm$0.67} & \textbf{47.08$\pm$0.67} & 38.94$\pm$0.61 & \textbf{50.25$\pm$0.63}
\\

\bottomrule
\end{tabular}
}
\end{table*}

\section{Better sampling improves cross-domain few-shot classification}
\label{app:cross_domain}

In~\cref{subsec:cross_domain}, we show that few-shot performance in the cross-domain setting can benefit from better sampling.
We train models on Mini-ImageNet (as done in~\cref{subsec:online}) and test the few-shot performance on the following datasets: CUB-200~\citep{welinder2010caltech}, Describable Textures~\citep{cimpoi2014describing}, FGVC-Aircraft~\citep{maji2013fine}, VGG Flowers~\citep{nilsback2006visual}.
We use \cnn and \resnet network architectures trained using ProtoNet (Euclidean and cosine), MAML and ANIL algorithms for the $5$-ways $1$- and $5$-shot settings.
Altogether, these makeup $64$ new scenarios.
We measure the accuracy on the test splits of~\citep{Triantafillou2020Meta-Dataset:}.

We compare online \uniform against baseline sampling and observe that online \uniform does statistically better in $49/64$ scenarios, comparable in $12/64$ scenarios, and worse in only $3/64$ scenarios.
The performance numbers are included in~\cref{tab:cross_domain}.

This further goes to show that sampling under the episodic training paradigm matters.
Using online \uniform leads to statistically significant improvements over the ubiquitous baseline sampling in most cases and rarely degrades performance.

\section{Number of trials}
\label{sec:num-trials}

In~\cref{tab:offline_full,tab:online_full} we make use of one random seed to give one training job per scenario per sampling method.
However we report performances over $1$k test episodes, as is typically done in few-shot learning.
We additionally ran $3$ training jobs for baseline sampling and online \uniform, resulting in $3$ training jobs per scenario per sampling method.
We observe that the difference in accuracy is $.20\%$ and $.02\%$ on average (ignoring the standard deviations) for baseline sampling and online \uniform; the effect of multiple random seeds is diminished when testing over many episodes.

\end{document}